\documentclass{osa-article}
\usepackage{subfigure}

\newcommand{\YKL}[1]{\textcolor{black}{#1}}
\newcommand{\CHANGE}[1]{\textcolor{black}{#1}}

\journal{oe}
\articletype{Research Article}
\begin{document}

\title{Polarization-driven Semantic Segmentation via Efficient Attention-bridged Fusion}
%\title{Deep Multimodal Semantic Segmentation with Single-Shot Polarization Perception and Efficient Attention-Bridged Fusion}

\author{Kaite Xiang,\authormark{1} Kailun Yang,\authormark{2} and Kaiwei Wang\authormark{1,3,*}}

\address{
\authormark{1}State Key Laboratory of Modern Optical Instrumentation, Zhejiang University, Hangzhou 310027, China\\
\authormark{2}Institute for Anthropomatics and Robotics, Karlsruhe Institute of Technology, 76131 Karlsruhe, Germany\\
\authormark{3}National Engineering Research Center of Optical Instrumentation, Zhejiang University, Hangzhou 310058, China\\
}

\email{\authormark{*}wangkaiwei@zju.edu.cn} %% email address is required



\begin{abstract}
Semantic Segmentation (SS) is promising for outdoor scene perception in safety-critical applications like autonomous vehicles, assisted navigation and so on.
However, traditional SS is primarily based on RGB images, which limits the reliability of SS in complex outdoor scenes, where RGB images lack necessary information dimensions to fully perceive unconstrained environments.
As preliminary investigation, we examine SS in an unexpected obstacle detection scenario, which demonstrates the necessity of multimodal fusion.
Thereby, in this work, we present EAFNet, an Efficient Attention-bridged Fusion Network to exploit complementary information coming from different optical sensors.
Specifically, we incorporate polarization sensing to obtain supplementary information, considering its optical characteristics for robust representation of diverse materials.
By using a single-shot polarization sensor, we build the first RGB-P dataset which consists of 394 annotated pixel-aligned RGB-Polarization images.
A comprehensive variety of experiments shows the effectiveness of EAFNet to fuse polarization and RGB information, as well as the flexibility to be adapted to other sensor combination scenarios.
\end{abstract}

%%%%%%%%%%%%%%%%%%%%%%%%%%  body  %%%%%%%%%%%%%%%%%%%%%%%%%%
\section{Introduction}
With the development of deep learning, outdoor scene perception and understanding has become a popular topic in the area of autonomous vehicles, navigation assistance systems for vulnerable road users like visually impaired pedestrians and mobile robotics~\cite{yang2018predicting}.
Semantic Segmentation (SS) is a task to assign semantic labels to each pixel of the images, \textit{i.e.}, object classification task at the pixel level, which is promising for outdoor perception applications~\cite{feng2020deep}.
A multitude of SS neural networks have been proposed following the trend of deep learning like FCN~\cite{long2015fully}, U-Net~\cite{ronneberger2015u}, ERFNet~\cite{romera2018erfnet}, SwiftNet~\cite{orvsic2019defense} and so on.

However, the networks mentioned above are mainly focused on the segmentation of RGB images, which makes it hard to fully perceive complex surrounding scenes because of the limited color information. 
A lot of works concerning domain adaptation have been presented to cope with SS in conditions without enough optical information~\cite{romera2019bridging,sun2019see}.
Yet, a high-level security needs to be guaranteed for outdoor scene perception to support safety-critical applications like autonomous vehicles, where merely algorithm advancement is insufficient.
Incorporating heterogeneous imaging techniques, multimodal semantic segmentation is of great necessity to be researched, which can leverage various optical information like depth, infrared and event-based data~\cite{zhang2020deep,zhang2020issafe}. 
In this paper, we employ polarization information as the supplement sensor information to advance the performance of RGB-based SS considering its optical characteristics for robust representation of diverse materials.
The polarization information are promising to advance the segmentation of objects which possess polarization features in the outdoor.
With the rationale, this work advocates polarization-driven multimodal SS, which is rarely explored in the literature.

\begin{figure}[t] 
  \centering 
  \subfigure[RGB Image]{ 
    \includegraphics[height=2cm]{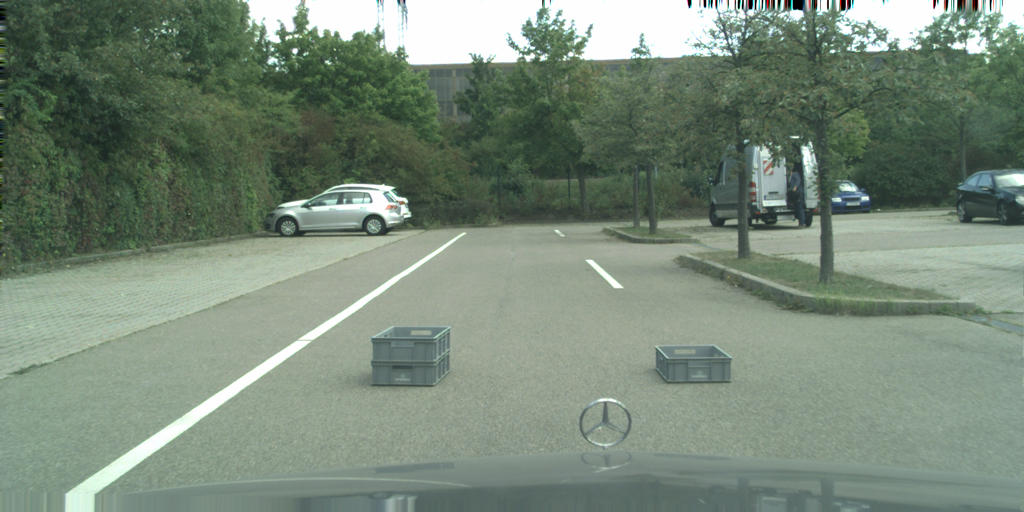}
  } 
  \subfigure[Disparity Image]{ 
    \includegraphics[height=2cm]{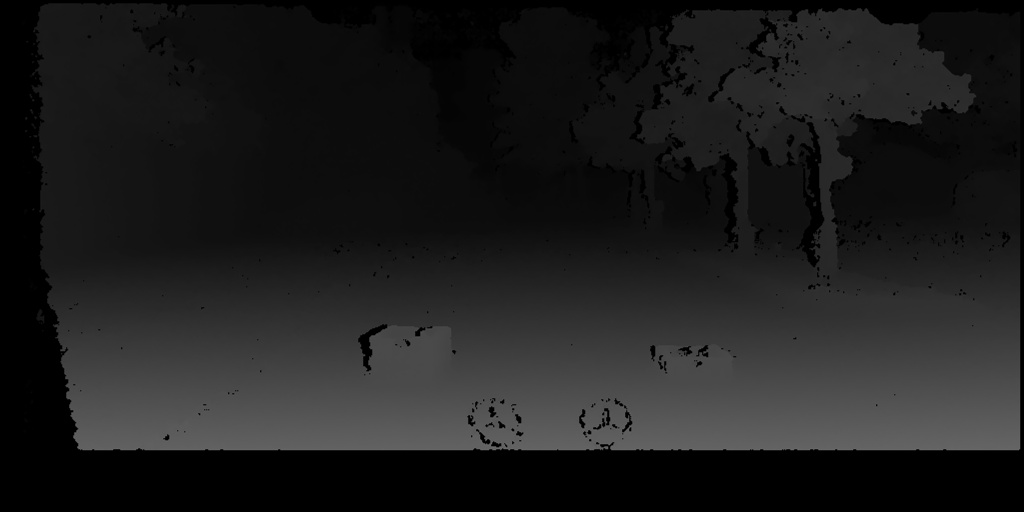} 
  } 
    \subfigure[Label]{ 
    \includegraphics[height=2cm]{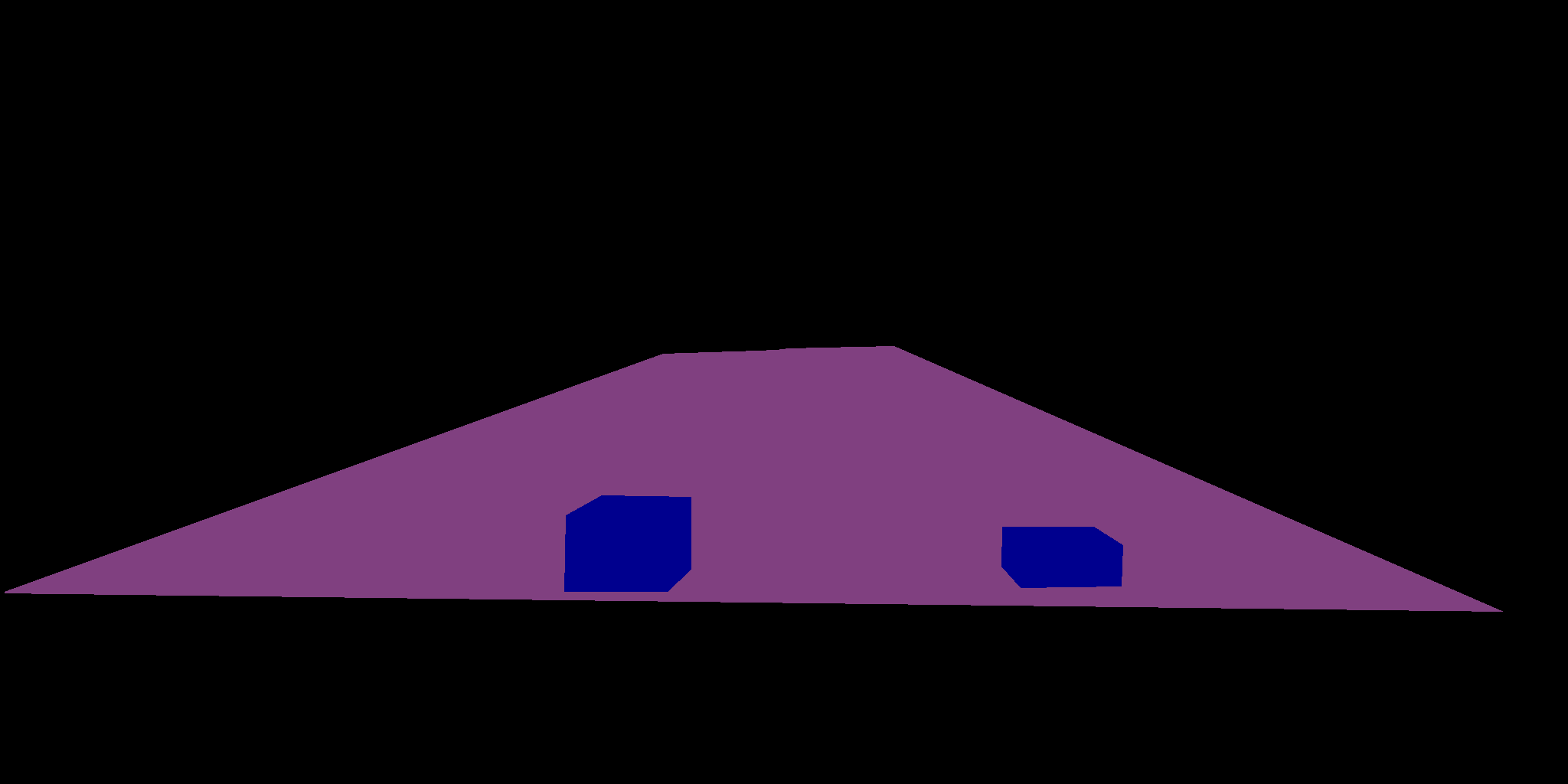} 
  } 
  \vskip-2ex
  \caption{\label{fig:1}An example of \textit{Lost and Found} dataset: (a) is the RGB image. (b) is the disparity image. (c) is the label, where the blue area denotes the obstacles, the purple area denotes the passable area and the black area denotes the background.} 
  \vskip-2ex
\end{figure}

\begin{figure}[t] 
  \centering 
  \subfigure[RGB Image]{ 
    \includegraphics[height=6cm]{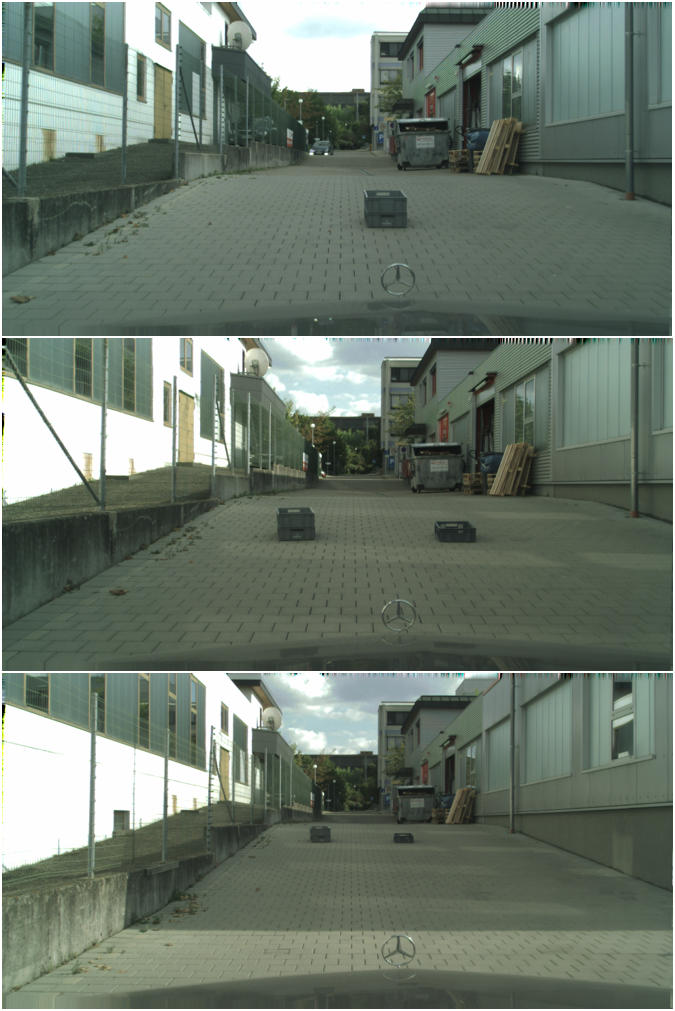}
  } 
  \subfigure[Label]{ 
    \includegraphics[height=6cm]{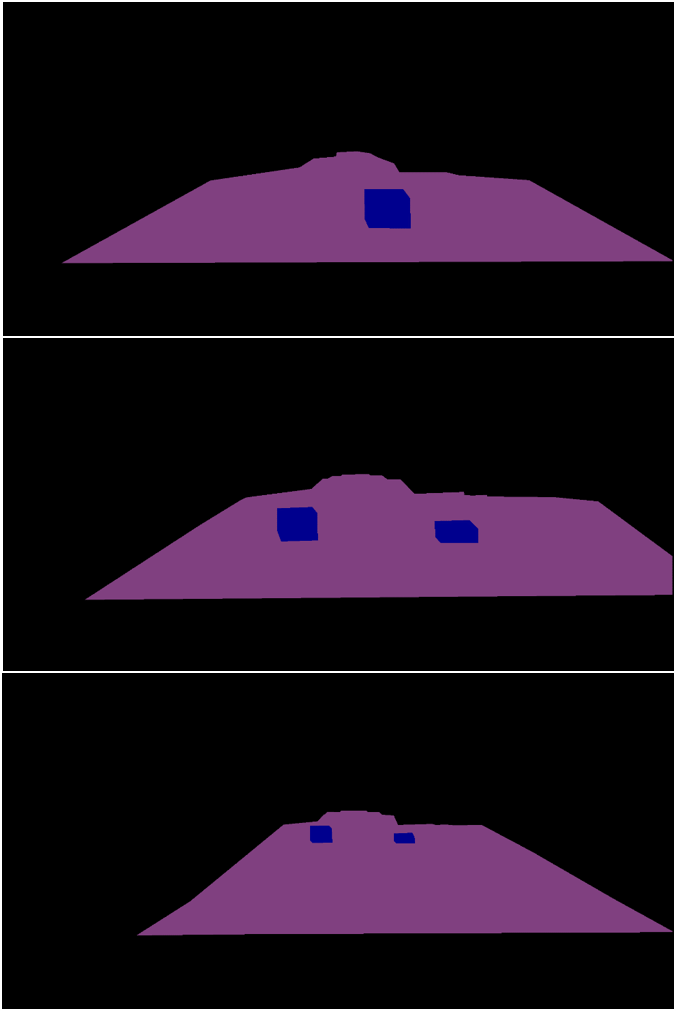} 
  } 
    \subfigure[Output]{ 
    \includegraphics[height=6cm]{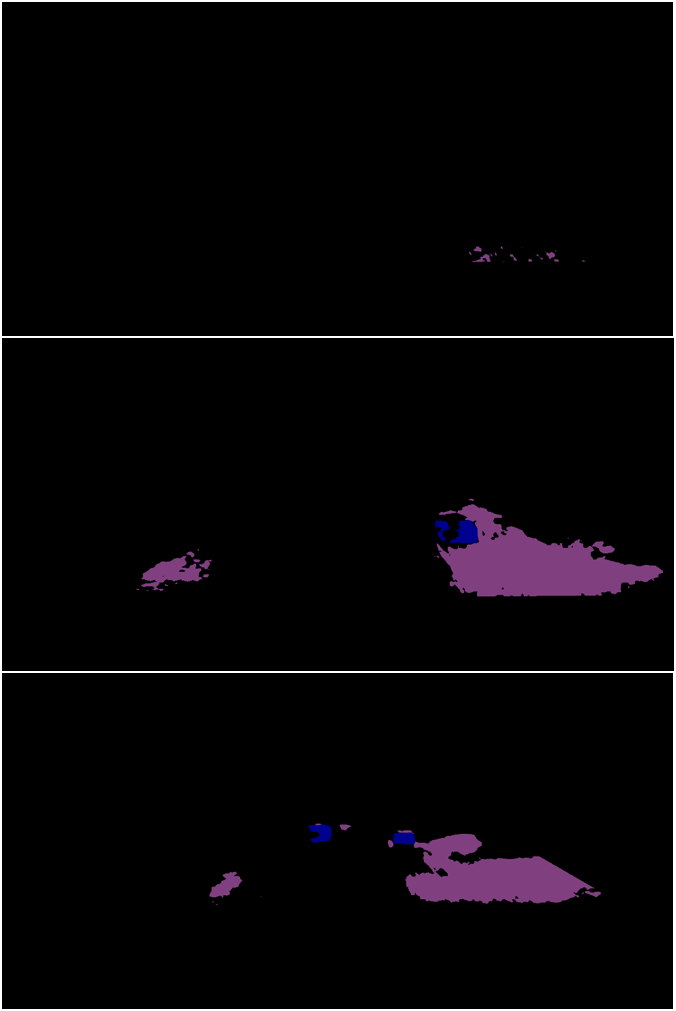} 
  } 
  \vskip-2ex
  \caption{\label{fig:2}The effect of SwiftNet~\cite{orvsic2019defense} trained merely with RGB images.} 
  \vskip-2ex
\end{figure}

To better explain the necessity of multimodal semantic segmentation that merely RGB sensors can not cope with complex outdoor scene perception,
we conduct a preliminary investigation in an unexpected obstacle detection scenario.
In outdoor scenes, many unexpected obstacles like tiny animals, boxes and so on are risk factors for secure driving.
We choose \textit{Lost and Found} dataset~\cite{pinggera2016lost} to perform an experiment.
The dataset is acquired by a pair of cameras with a baseline distance of 23cm in 13 challenging outdoor traffic scenes by setting up 37 different categories of tiny obstacles, which possesses three types of data as shown in Fig.~\ref{fig:1}, \textit{i.e.}, RGB image, disparity image and ground-truth label.
The dataset contains 3 categories, \textit{i.e.}, coarse annotations of passable areas, fine-grained annotations of unexpected tiny obstacles and background, whose resolution is 1024$\times$2048.
Among them, 1036 images are selected as the training set, while the remaining 1068 images are selected as the validation set.
Considering the fact that outdoor scene perception application demands high efficiency, we select a real-time network SwiftNet~\cite{orvsic2019defense} to conduct the experiment.
We only take the RGB images as the input information to train the network, where other training implementations will be described in Section 4.1.
The detailed results are shown in Table~\ref{tab:3}.
The precision of obstacle and passable area segmentation are 26.4\% and 85.4\%. Their recall rate are 49.8\% and 63.9\%, and their Intersection over Union (IoU) are 20.9\% and 56.2\%, respectively.
In addition, the qualitative results of the experiment are illustrated in Fig.~\ref{fig:2}.
The results show that severe over-fitting has appeared, and we find that the model trained merely with RGB images can not satisfactorily detect small, unexpected obstacles.

According to the toy experiment above, the model's performance is unacceptable when trained only with RGB images.
Thereby, we consider it is necessary to incorporate additional sensor information for semantic segmentation to perceive outdoor traffic scenes.
As mentioned above, we select polarization as the complementary information, whose potential has been shown in our previous works~\cite{yang2017detecting,sun2019multimodal} for water hazard detection.
In this work, we leverage a novel single-shot RGB-P imaging sensor, and investigate polarization-driven semantic segmentation.
To sufficiently fuse RGB-P information, we propose the Efficient Attention-bridged Fusion Network (EAFNet), enabling adaptive interaction of cross-modal features. In summary, we deliver the following contributions:  
\begin{itemize}
    \item Addressing polarization-driven semantic segmentation, we propose EAFNet, an efficient attention-bridged fusion network, which fuses multimodal sensor information with a lightweight fusion module, advancing many categories' accuracy, especially categories with polarization characteristics like glass, whose IoU is lifted to 79.3\% from 73.4\%. The implementations and codes will be made available at \url{https://github.com/Katexiang/EAFNet}.
    \item With a single-shot polarization imaging sensor, we present an RGB-P outdoor semantic segmentation dataset. To the best of our knowledge, this is the first RGB-P outdoor semantic segmentation dataset, which will be made publicly accessible at \url{http://www.wangkaiwei.org/download.html}. 
    \item We conduct a series of experiments to demonstrate the effectiveness of EAFNet with comprehensive analysis, along with a supplementary experiment that verifies EAFNet's generalization capability for fusing other sensing data besides polarization information.
\end{itemize}

\section{Related work}

\subsection{From accurate to efficient semantic segmentation}

Convolutional Neural Networks (CNNs) have been the mainstream solution to semantic segmentation since Fully Convolutional Networks (FCNs)~\cite{long2015fully} approached the dense recognition task in an end-to-end way.
SegNet~\cite{badrinarayanan2017segnet} and U-Net~\cite{ronneberger2015u} presented encoder-decoder architectures, which are widely used in the following networks.
Benefiting from deep classification models like ResNets~\cite{he2016deep}, PSPNet~\cite{zhao2017pyramid} and DeepLab~\cite{chen2017deeplab} constructed multi-scale representations and achieved significant accuracy improvements.
Inspired by the channel attention method proposed in SENet~\cite{hu2018squeeze}, EncNet~\cite{zhang2018context} encoded global image statistics, while HANet~\cite{choi2020cars} explored height-driven contextual priors.
ACNet~\cite{hu2019acnet} leveraged attention connections and bridged multi-branch ResNets to exploit complementary features.
In another line, DANet~\cite{fu2019dual} and OCNet~\cite{yuan2018ocnet} aggregated dense pixel-pair associations.
These works have pushed the boundary of segmentation accuracy and attained excellent performances on existing benchmarks.

In addition to accuracy, the efficiency of segmentation CNNs is crucial for real-time applications.
Efficient networks were designed such as ERFNet~\cite{romera2018erfnet} and SwiftNet~\cite{orvsic2019defense}.
They were built on techniques including early downsampling, filter factorization, multi-branch setup and ladder-style upsampling.
Some efficient CNNs~\cite{yang2019ds,sun2020real} also leveraged attention connections, trying to improve the trade-off between segmentation accuracy and computation complexity.
With these advances, semantic segmentation can be performed both swiftly and accurately, and thereby has been incorporated into many optical sensing applications such as semantic cognition system~\cite{yang2019robustifying} and semantic visual odometry~\cite{yang2019ds,chen2019palvo}.

\subsection{From RGB-based to multimodal semantic segmentation}

While ground-breaking network architectural advances have been achieved in single RGB-based semantic segmentation on existing RGB image segmentation benchmarks such as Cityscapes~\cite{cordts2016cityscapes}, BDD~\cite{yu2020bdd100k} and Mapillary Vistas~\cite{neuhold2017mapillary}, in some complex environments or under challenging conditions, it is necessary to employ multiple sensing modalities that provide complementary information of the same scene.
Comprehensive surveys on multimodal semantic segmentation were presented in~\cite{feng2020deep,zhang2020deep}.
In the literature, researchers explored RGB-Depth~\cite{hu2019acnet,sun2020real}, RGB-Infrared~\cite{valada2016deep,choe2018ranus}, RGB-Thermal~\cite{li2020segmenting,vertens2020heatnet}, GRAY-Polarization~\cite{zhang2019exploration,blanchon2019outdoor} and Event-based~\cite{zhang2020issafe,alonso2019ev}
semantic segmentation to improve the reliability of surrounding sensing and the applicability towards real-world applications.
For example, RFNet~\cite{sun2020real} fused RGB-D information on heterogeneous datasets, improving the robustness of SS in road-driving scenes with small-scale, unexpected obstacles.

In this work, we focus on RGB-P semantic segmentation by using a single-shot polarization camera.
Traditional polarization-driven dense prediction frameworks were mainly dedicated to the detection of water hazards~\cite{yang2018perception,han2018single} or the perception in indoor scenes~\cite{huang2017target,berger2017depth}.
In our previous works, we investigated the impact of loss functions on water hazard segmentation~\cite{xiang2019importance}, followed by a comparative study on high-recall semantic segmentation~\cite{xiang2019comparative}.
Inspired by~\cite{huang2017target}, dense polarization maps were predicted from RGB images through deep learning~\cite{yang2018predicting}.
Instead, current polarization imaging technique makes it possible to sense pixel-wise polarimetric information in a single shot and has been integrated on perception platforms for autonomous vehicles~\cite{sun2019multimodal}.
Following this line, we present a multimodal semantic segmentation system with single-shot polarization sensing.
Notably, we found previous collections~\cite{wang2017multimodality,zhang2019exploration,blanchon2019outdoor} of polarization images were mainly gray images without providing RGB information that are critical for segmentation tasks.
Besides, they were limited in terms of data diversity and entailed careful calibration between different cameras.
In contrast, we are able to bypass the complex calibration and naturally obtain multimodal data with single-shot polarization imaging.
As an important contribution of this work, a novel outdoor traffic scene RGB-P dataset is collected and densely annotated, which covers not only specular scenes but also diverse unstructured surroundings.
The dataset will be made publicly available to the community to foster polarimetry-based semantic segmentation.
Moreover, our work is related to transparent object segmentation~\cite{xie2020segmenting,kalra2020deep}.

\YKL{In addition, some polarization-driven objection detection methods have been explored in~\cite{blin2019road,wang2020end}.
The designed modules re-encode raw polarization images for a better representation of polarization information. It may be beneficial when the input images are all of the same type like polarization raw images. After all, they are RGB images and possess similar data distribution. However, it can not be generalized into other sensors easily like RGB and disparity images, which have a distinct difference in data distribution. Unlike them, our work is to build a network architecture which can be flexibly adapted to different sensors besides RGB-polarization information.
}

\section{Methodology}
In this section, we derive the polarization image formation process and explain why polarization images contain rich information to complement RGB images for semantic segmentation.
Then, we make a brief introduction of our integrated multimodal sensor and the novel RGB-P dataset.
Finally, we present the Efficient Attention-bridged Fusion Network (EAFNet) for polarimetry-based multimodal scene perception.

   \begin{figure} [t]
   \begin{center}
   \begin{tabular}{c} %% tabular useful for creating an array of images 
   \includegraphics[height=4cm]{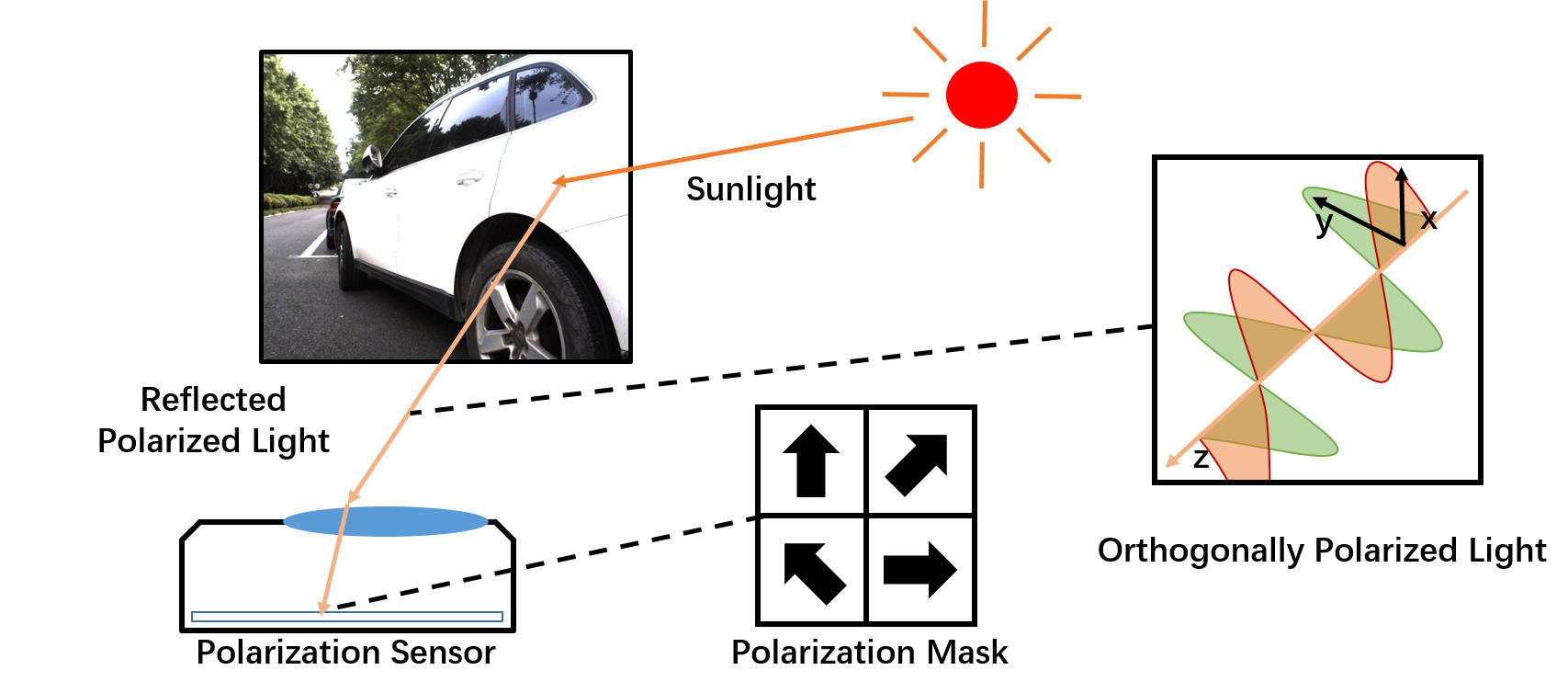}
   \end{tabular}
   \end{center}
   \vskip-4ex
   \caption[example] 
   { \label{fig:3}Polarization image formation model.}
   \vskip-2ex
   \end{figure}
   
\subsection{Polarization image formation}

\YKL{Polarization is a significant characteristic of electromagnetic waves. When optical flux is incident upon a surface or medium, three processes occur: reflection, absorption and transmission. Analyzing the polarization of reflected light is possible to determine the optical proprieties of a given surface or medium.}
We illustrate the importance of polarization according to the Fresnel equation:
\begin{equation}
\begin{array}{l}
r_{s}=\frac{n_{1} \cos \theta_{i}-n_{2} \cos \theta_{t}}{n_{1} \cos \theta_{i}+n_{2} \cos \theta_{t}} \, ,\quad t_{s}=\frac{2 n_{1} \cos \theta_{i}}{n_{1} \cos \theta_{i}+n_{2} \cos \theta_{t}} \, ,\ \\
r_{p}=\frac{n_{2} \cos \theta_{i}-n_{1} \cos \theta_{t}}{n_{2} \cos \theta_{i}+n_{1} \cos \theta_{t}} \, ,\quad t_{p}=\frac{2 n_{1} \cos \theta_{i}}{n_{2} \cos \theta_{i}+n_{1} \cos \theta_{t}} \, ,\
\end{array}
\end{equation}
where \textit{r$_{s}$} and \textit{t$_{s}$} are the reflected and refracted portion of incoming light, the subscript label \textit{s} and \textit{p} represent perpendicular polarization and parallel polarization, \textit{n$_{1}$} and \textit{n$_{2}$} are refractive indexes of the two media material, and the \textit{$\theta_{i}$} and \textit{$\theta_{t}$} are the angle of incident light and refracted light, respectively.
Inferred from Eq. (1), we find that the surface material's optical characteristics can affect the intensity of the two orthogonally polarized light.
Therefore, the orthogonally polarized light can partially reflect the surface material. 

The polarization image formation can be reducible to the model shown in Fig.~\ref{fig:3}.
In outdoor scenes, the light source is mainly sunlight.
When the sunlight shines on the object like cars, polarized reflection occurs.
Then, the reflected light with orthogonally polarized portion enters the camera with a polarization sensor, and the optical information with polarized characteristics are recorded by the sensor.
The reason why the photoelectric sensor can record the polarized information is that the sensor's surface is covered by a polarization mask layer with four different polarization directions, and only the light with the same polarization direction can pass the layer.
\begin{figure}[t] 
  \centering 
  \subfigure[RGB images with four polarization directions]{ 
    \includegraphics[height=5cm]{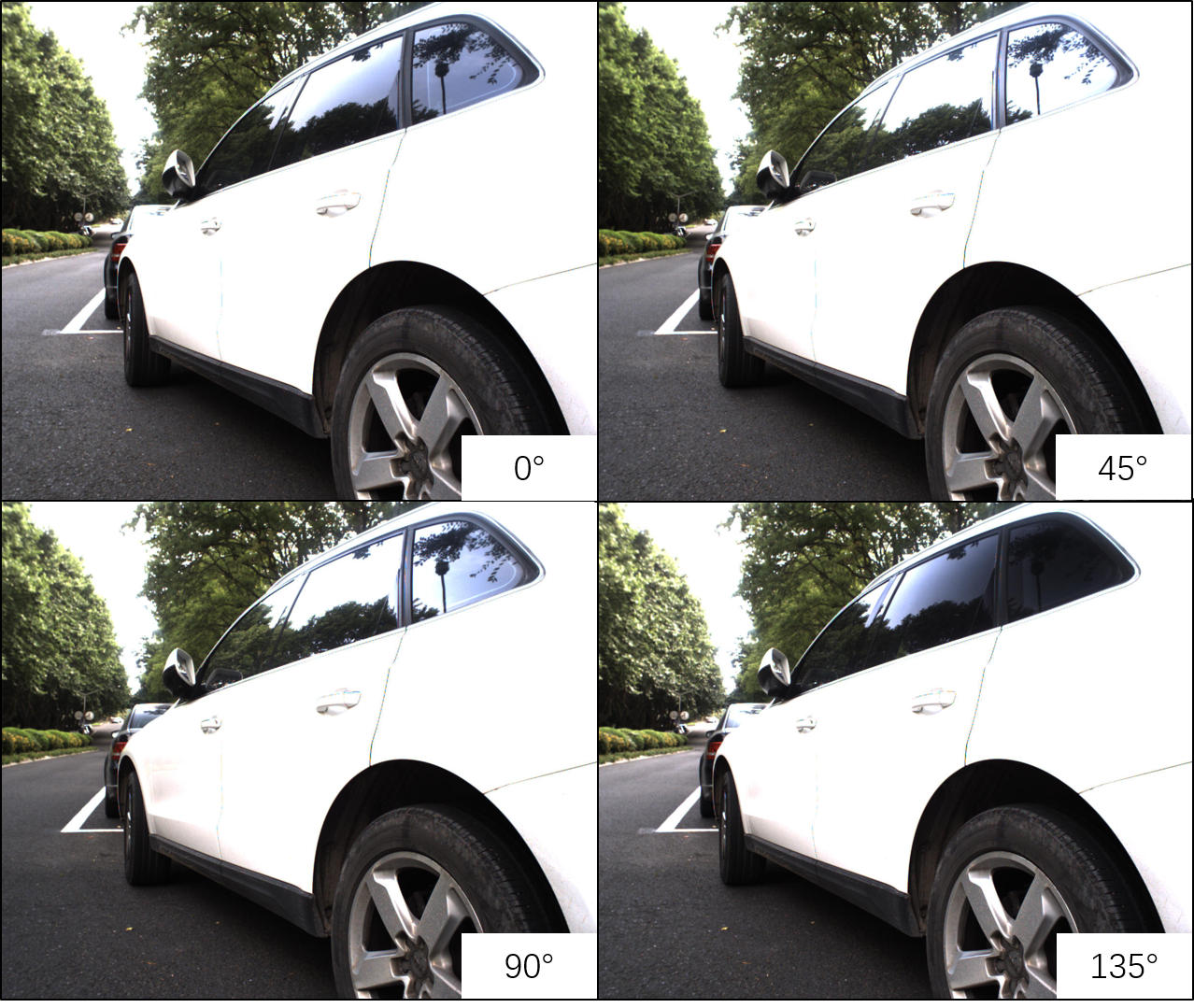}
  } 
  \subfigure[Polarization image]{ 
    \includegraphics[height=5cm]{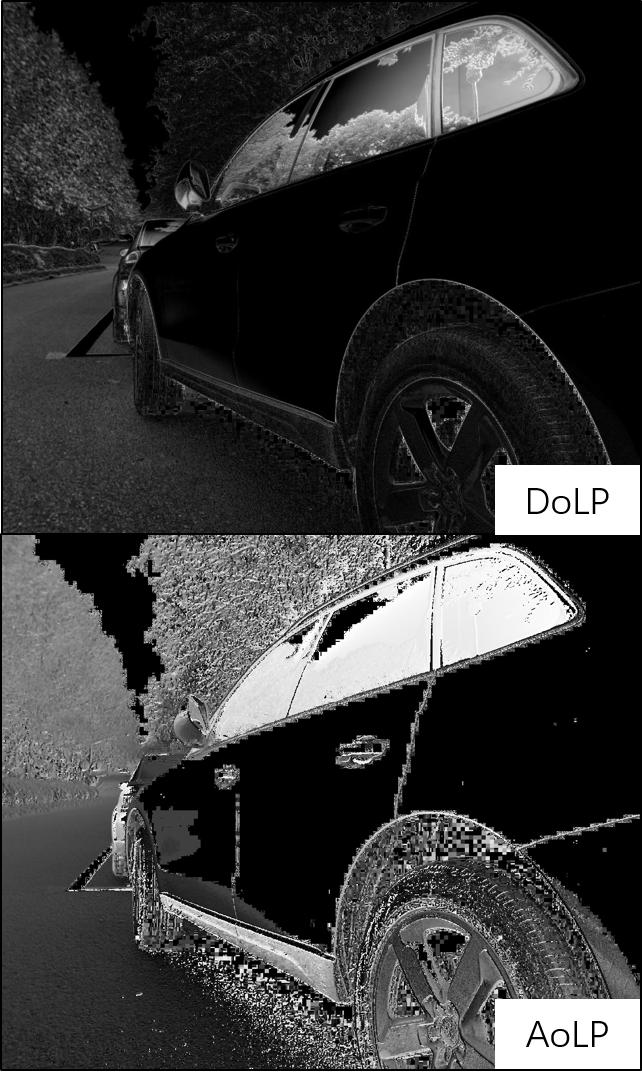} 
  } 
  \vskip-2ex
  \caption{\label{fig:4}Display of polarization characteristics: (a) RGB images at four polarization directions; (b) Polarization image.} 
  \vskip-2ex
\end{figure}   

Here, we make a brief introduction of polarization parameters like the Degree of Linear Polarization (DoLP) and the Angle of Linear Polarization (AoLP).
They are the key elements that contribute to the advancement of multimodal semantic segmentation.
They are derived by Stokes vectors \textbf{\textit{S}}, which are composed of four parameters, \textit{i.e.}, \textbf{\textit{S$_{0}$}}, \textbf{\textit{S$_{1}$}}, \textbf{\textit{S$_{2}$}} and \textbf{\textit{S$_{3}$}}.
More precisely, \textbf{\textit{S$_{0}$}} stands for the total light intensity, \textbf{\textit{S$_{1}$}} stands for the parallel polarized portion's superiority against the perpendicular polarized portion, and \textbf{\textit{S$_{2}$}} stands for 45$^{\circ}$ polarized portion's superiority against 135$^{\circ}$ polarized portion.
\textbf{\textit{S$_{3}$}}, associated to circularly polarized light, is not involved in our work on multimodal semantic segmentation.
They can be derived by:
\begin{equation}
\left\{\begin{array}{l}
S_{0}=I_{0}+I_{90}=I_{45}+I_{135}  \, ,\ \\
S_{1}=I_{0}-I_{90}  \, ,\ \\ 
S_{2}=I_{45}-I_{135} \, ,\
\end{array}\right.
\end{equation}
where \textit{I$_{0}$}, \textit{I$_{45}$}, \textit{I$_{90}$} and \textit{I$_{135}$} are the optical intensity values at the certain polarization direction, \textit{i.e.}, 0$^{\circ}$, 45$^{\circ}$, 90$^{\circ}$ and 135$^{\circ}$.
Here, DoLP and AoLP can be formulated as:
\begin{equation}
D o L P=\frac{\sqrt{S_{1}^{2}+S_{2}^{2}}}{S_{0}} \, ,\
\end{equation}
\begin{equation}
A o L P=\frac{1}{2} \times \arctan \left(\frac{S_{1}}{S_{2}}\right) .
\end{equation}
According to Eq. (3), the range of DoLP is from 0 to 1.
For partially polarized light, DoLP $\in(0,1)$.
For completely polarized light, DoLP $=1$.
Namely, DoLP stands for the degree of Linear Polarization.
For AoLP, it ranges from 0$^{\circ}$ to 180$^{\circ}$. 
AoLP can reflect object's silhouette information, because objects with the same material normally possess similar AoLP.
Therefore, AoLP is a natural scene segmentation mask.
In other words, objects of the same category or with the same material have similar AoLP.
\CHANGE{Different from RGB-based sensors whose output can be influenced by various outdoor environments like foggy weather or dust, polarization information from RGB-P sensor still keeps stable according to Eq. (2), (3) and (4). Because the RGB images with four polarization directions from RGB-P sensor suffer similar degradation in the process of reaching the polarization filter in RGB sensor and the degradation will be cancelled out in the derivation of DoLP and AoLP. Therefore, the polarization stability against various outdoor environments is beneficial for scene perception.}
We generate a visualization of a set of DoLP and AoLP polarization images, as shown in Fig.~\ref{fig:4}.
We find that the glass area and vegetation area are of high DoLP,
but other areas are of low DoLP,
which offers limited information, merely focused on the area with polarized characteristics. 
Besides, the left part of the vegetation and sky can not be distinguished merely depending on DoLP. 
On the contrary, for AoLP, we find areas of the same category show proper continuity of polarization information, which indicates great spatial priors for SS.
AoLP offers a better representation of spatial information, which keeps a consistent distribution on the same category or materials like vegetation, sky, road and glass.
In Section 4, we will further analyze the AoLP's great potential in providing extra spatial information for SS over DoLP with extensive experiments.

\subsection{Integrated multimodal sensor and ZJU-RGB-P dataset}
\begin{figure}[t] 
  \centering 
  \subfigure[Integrated multimodal sensor]{ 
    \includegraphics[height=3.5cm]{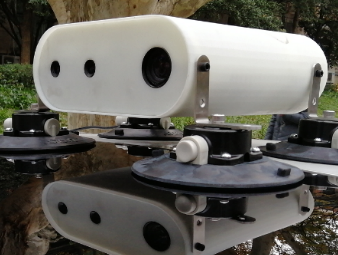}
  } 
  \subfigure[Output of the sensor]{ 
    \includegraphics[height=3.5cm]{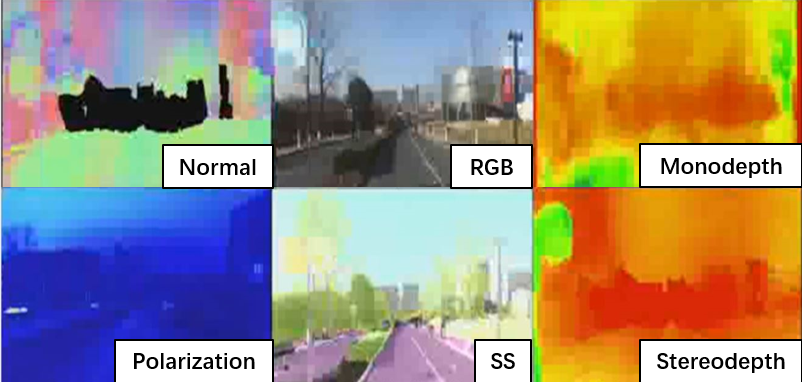} 
  } 
  \vskip-2ex
  \caption{\label{fig:5}Our integrated multimodal vision sensor for capturing polarization information. (a) The integrated multimodal sensor; (b) The output of the sensor.} 
  \vskip-2ex
\end{figure}

The RGB-P outdoor scene dataset is captured by an integrated multimodal sensor for autonomous driving~\cite{sun2019multimodal}, as shown in Fig.~\ref{fig:5}. 
The sensor is a highly integrated system which is a combination of polarization sensor, RGB sensor, infrared sensor and depth sensor.
The sensor captures polarization information with an RGB-based imaging sensor \textbf{LUCID\_PHX050S}. The difference between \textbf{LUCID\_PHX050S} and Gray-Polarization sensors is that the former is covered with an extra Bayer array besides the polarization mask.
\CHANGE{Intensity and luminance are important for the detection quality with image sensors,
but the use of polarization (eg., linear polarization ) filters naturally results in losing less of half of input intensity coming to the RGB sensor.
The reason why the collected RGB-polarization images can keep the consistency against the changing outdoor environment is that we utilize the data acquisition and processing program, \textit{i.e.}, \textbf{Arena}, that can dynamically adjust the gain factor and exposure time according to the outdoor environment.}
In addition, the multimodal sensor integrates an embedded system which combines hardware and software, by which we can attain various types of information like semantic information, infrared information, \YKL{stereo depth information~\cite{hirschmuller2005accurate}, monocular depth information~\cite{zhou2020robust} and surface normal information~\cite{li2015depth} by utilizing relevant estimation algorithms. While the sensor provides diverse modalities, this work focuses on using RGB and polarization information.} 
Fig.~\ref{fig:5}(b) shows some examples of the sensor's output information. 
The highly integrated sensor can broaden RGB-based sensor's application scenarios~\cite{sun2019multimodal}.
The infrared information can assist nighttime semantic segmentation, and the polarization-RGB-infrared multimodal sensor can offer precise depth information by pairing the sensors with different baselines.
We leverage the multimodal sensor to attain pixel-aligned polarization and RGB images, and the main purpose of this work to adapt RGB-based SS to Polarization-driven multimodal SS.

RGB-Polarization outdoor scene SS dateset is scarce in the literature.
Some research groups have realized the importance of polarization information for outdoor perception.
The Polabot dataset~\cite{zhang2019exploration,blanchon2019outdoor}, a GARY-Polarization outdoor scene SS dataset, consists of around 180 pairs of images at a low resolution of 230$\times$320.
The limited image number and the low resolution make it hard to train a robust SS network for outdoor scenes.
In addition, the dataset is short of RGB information which provide important texture features for classification tasks.

\begin{figure} [t]
   \begin{center}
   \begin{tabular}{c} %% tabular useful for creating an array of images 
   \includegraphics[height=2.5cm]{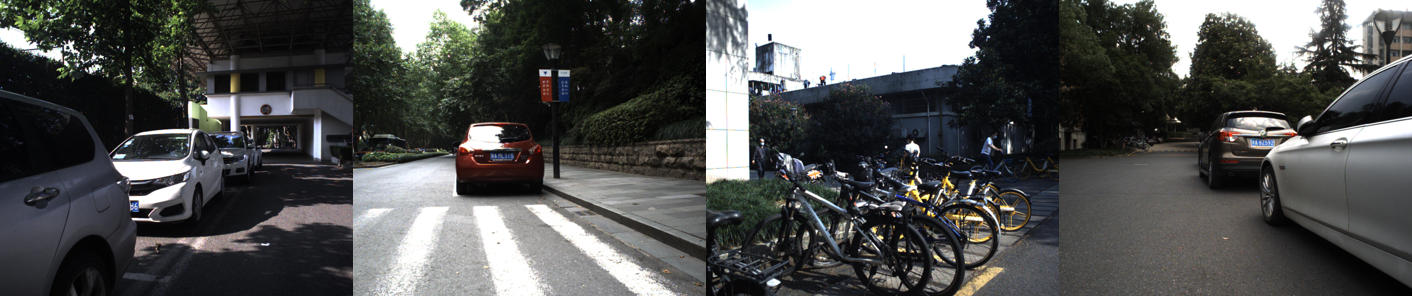}
   \end{tabular}
   \end{center}
   \vskip-4ex
   \caption[example] 
   { \label{fig:6}Diverse scenes in our ZJU-RGB-P dataset.}
   %\vskip-2ex
   \end{figure}

Addressing the scarcity, we build the first RGB-P outdoor scene dataset which consists of 394 annotated pixel-aligned RGB-Polarization images.
We collect the images with abundant and complex scenes at Yuquan Campus, Zhejiang University, as shown in Fig.~\ref{fig:6}.
The scenes of the dataset cover road scenes around teaching building area, canteen area, library and so on to provide diverse scenes for reducing the risk of over-fitting in training SS models.
   
    \begin{figure} [t]
   \begin{center}
   \begin{tabular}{c} %% tabular useful for creating an array of images 
   \includegraphics[height=6cm]{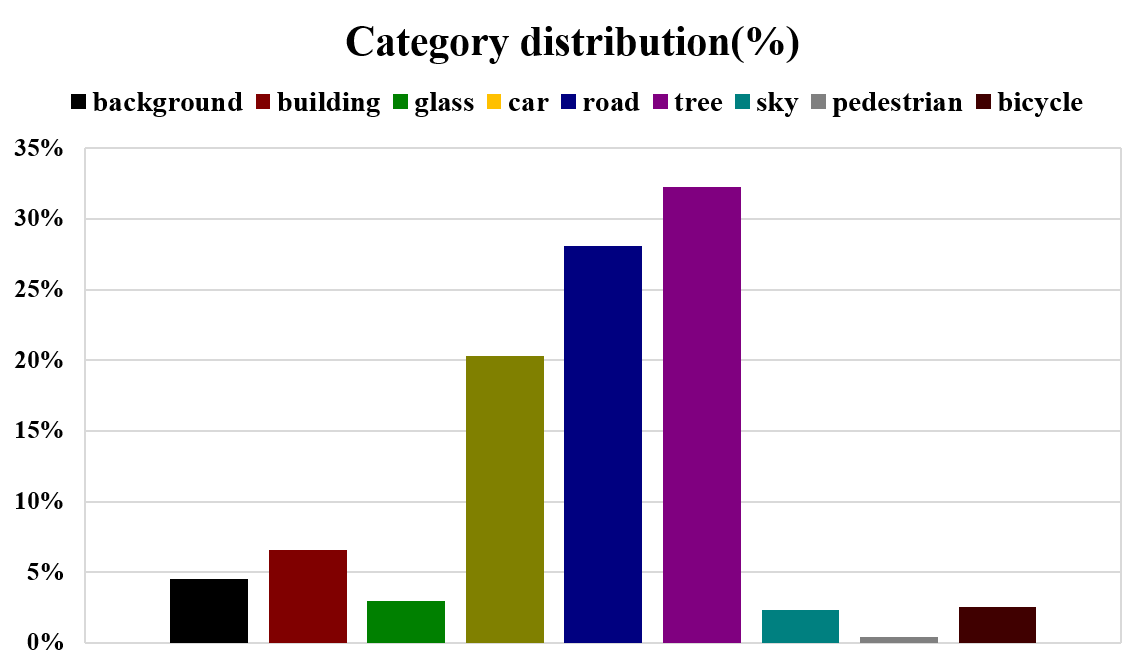}
   \end{tabular}
   \end{center}
   \vskip-4ex
   \caption[example] 
   { \label{fig:6.1}Histogram of ZJU-RGB-P dataset's category distribution.}
   %\vskip-2ex
   \end{figure}  
   
   \begin{figure} [t]
   \begin{center}
   \begin{tabular}{c} %% tabular useful for creating an array of images 
   \includegraphics[height=4cm]{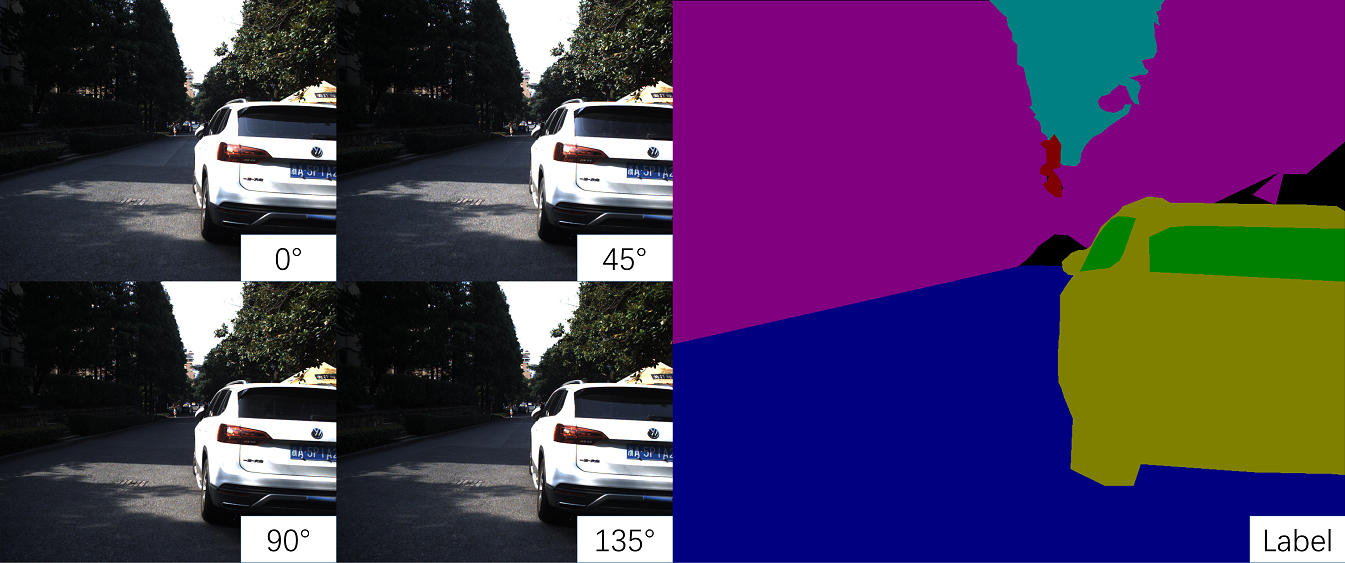}
   \end{tabular}
   \end{center}
   \vskip-4ex
   \caption[example] 
   { \label{fig:7}An example of ZJU-RGB-P dateset including RGB images at different polarization directions and the pixel-wise semantic segmentation label.}
   \vskip-2ex
   \end{figure}
   
The resolution of our dataset is 1024$\times$1224, which makes it possible to apply data augmentation like random crop and random rescale, which are crucial for improving data diversity and attaining robust segmentation~\cite{yang2019robustifying}.
%Making an analysis of the category distribution,
We label the dateset with 9 classes at the pixel level, \textit{i.e.}, Building, Glass, Car, Road, Vegetation, Pedestrian, Bicycle and Background \YKL{using LabelMe~\cite{russell2008labelme}. Here, we make a statistics of category distribution at the pixel level and draw a histogram as shown in Fig.~\ref{fig:6.1}. We can learn from the histogram that the dataset has a diversity of categories, and the categories with a low pixel proportion will be the difficult categories for SS like glass and pedestrian.}  
An example of the dataset is shown in Fig.~\ref{fig:7}, which consists of four pixel-aligned RGB images at four polarization directions and a SS label.
But AoLP and DoLP are the ultimate polarization representations integrated into SS, so the four polarized RGB images need to be operated according to  Eq. (3) and  Eq. (4).
\YKL{We utilize the average of the four polarized RGB images as the RGB images to feed into EAFNet. In fact, the average of any two orthogonal directions RGB images can represent images captured by a conventional RGB sensor.} 
Finally, we select 344 images as the training set, and the other 50 images as the validation set. We name it \textit{ZJU-RGB-P} dataset.

\subsection{Efficient attention-bridged fusion network}
\begin{figure} [t]
   \begin{center}
   \begin{tabular}{c} %% tabular useful for creating an array of images 
   \includegraphics[height=5.2cm]{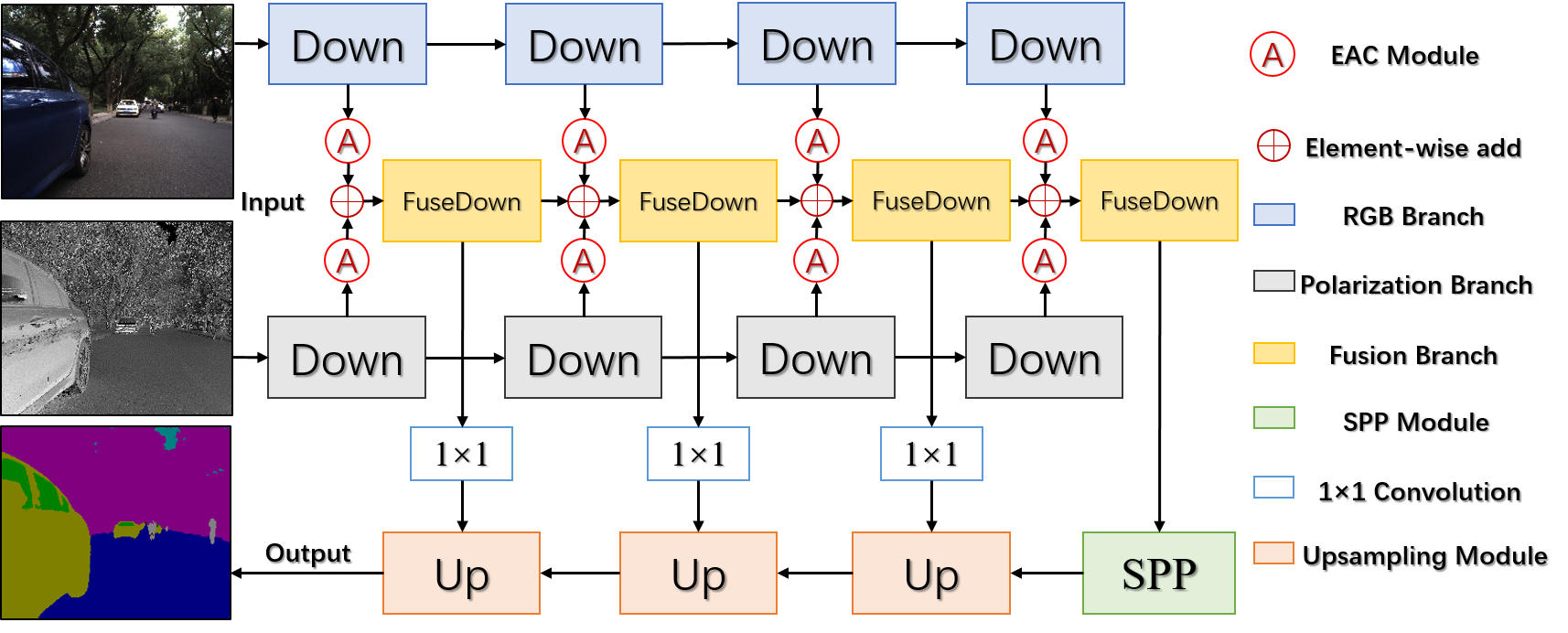}
   \end{tabular}
   \end{center}
   \vskip-4ex
   \caption[example] 
   { \label{fig:8}Overview of EAFNet. RGB and polarization images are input to the network for extracting features separately. The EAC modules adaptively fuse the features.}
   %\vskip-2ex
   \end{figure}

In order to combine RGB and polarization features, we present EAFNet, an Efficient Attention-bridged Fusion Network to exploit multimodal complementary information, whose architecture is shown in Fig.~\ref{fig:8}. Inspired by SwiftNet~\cite{orvsic2019defense} and our previous SFN~\cite{xiang2019comparative} with an U-shape encoder-decoder structure, EAFNet is designed to keep a similar architecture with downsampling paths to extract features and an upsampling module to restore the resolution, together with EAC modules to fuse features from RGB and polarization images.
Here, we make a brief overview of EAFNet according to Fig.~\ref{fig:8}.
%All the modules of EACNet are of the same as SwiftNet except the fusion module.
EAFNet is designed to have a three-branch structure with downsampling paths of the same type.
They are the RGB branch, the polarization branch and the fusion branch.
In order to advance computation efficiency, we employ ResNet-18~\cite{he2016deep}, a light-weight encoder to extract and fuse features.
After attaining the downsampled and fused features, SPP module, a spatial pyramid pooling module~\cite{he2015spatial,orvsic2019defense} is leveraged to enlarge valid receptive field.
Then, a series of upsampling modules are leveraged to restore the feature resolution.
Like SwiftNet, EAFNet employs a series of convolution layers with kernel size of 1$\times$1 to connect features between shallow layers and deep layers.
The key innovation, here, lies in that EAFNet possesses the carefully designed fusion module, namely EAC Module, with inspiration gathered from the Efficient Channel Attention Network~\cite{wang2020eca}.
With this architecture, EAFNet is a real-time network, whose inference speed on GTX 1080Ti reaches 24 FPS (Frame Per Second) at the resolution of 512$\times$1024. 
   
\begin{figure} [t]
   \begin{center}
   \begin{tabular}{c} %% tabular useful for creating an array of images 
   \includegraphics[height=3cm]{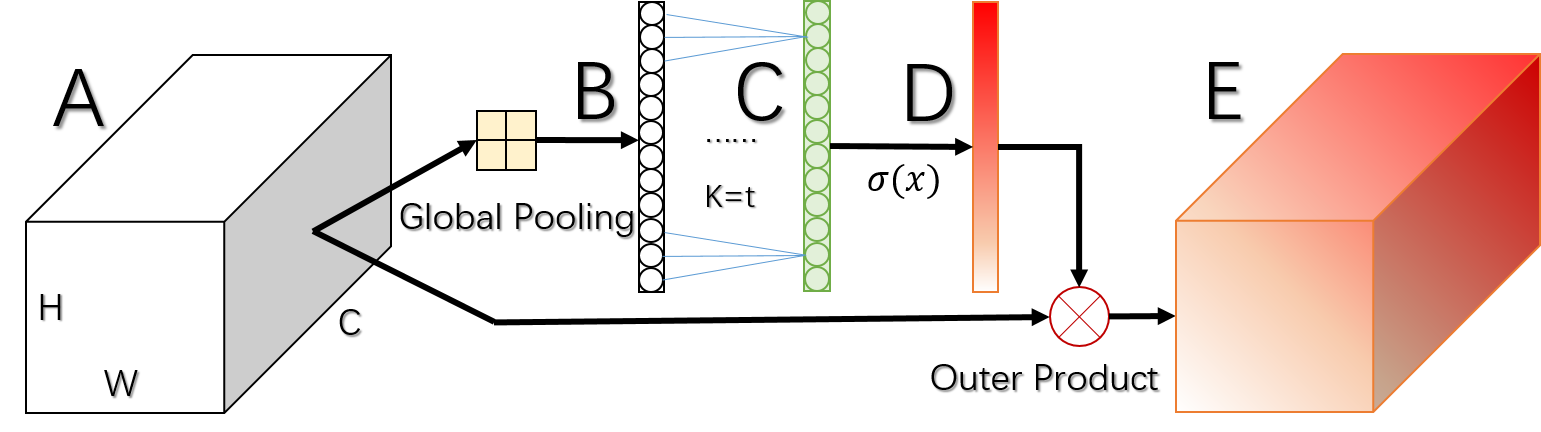}
   \end{tabular}
   \end{center}
   \vskip-4ex
   \caption[example] 
   { \label{fig:9}The Efficient Attention Complementary module (EAC module). A is the input feature map, B is the average global feature vector, C is the feature vector after a convolution layer with an adaptive kernel size. D is the attention weights, \textit{i.e.} the vector after activation function of C, and E is the adjusted feature map.}
   \vskip-2ex
   \end{figure}

\textbf{EAC Module} is an efficient attention complementary module which is designed for extracting informative RGB features and polarization features, as shown in Fig.~\ref{fig:9}.
It is an efficient version of the Attention Complementary Module (ACM)~\cite{hu2019acnet}, which replaces fully connection layers with 1$\times$1 convolution layers whose kernel sizes are adaptively determined according to the channel number of the corresponding feature maps.
On the one hand, the structure reduces computation complexity compared with ACM due to the use of local cross-channel interactions other than all channel-pair interactions.
On the other hand, local cross-channel interaction effectively avoids the problem of losing information caused by the dimension reduction in learning channel attention.

Assuming the input feature map is $ \boldsymbol{A} \in \mathbb{R}^{H \times W \times C}$, we first apply a global average pooling layer to process $ \boldsymbol{A}$, where $ \boldsymbol{H}$, $\boldsymbol{W}$ and $ \boldsymbol{C}$ are the height, width and channel number of the input feature map, respectively.
Then, we obtain a feature vector $B=\left[B_{1}, B_{2}, \ldots \ldots, B_{C}\right] \in \mathbb{R}^{1 \times C}$, where the subscript label represents the sequence number of features' channel.
The \textit{k}-th ($k \in[1, C]$) element of $ \boldsymbol{B}$ can be expressed as:
\begin{equation}
B_{k}=\frac{1}{H \times W} \sum_{i=1}^{H} \sum_{j=1}^{W} A_{(i, j)}^{k} .
\end{equation}
Then, the vector $ \boldsymbol{B}$ needs to be reorganized by a convolution layer with an adaptive kernel size \textbf{K} to obtain a more meaningful vector $C=\left[C_{1}, C_{2}, \ldots \ldots, C_{C}\right] \in \mathbb{R}^{1 \times C}$.
\textbf{K} is the key point to attain the local cross-channel interaction attention weights, which can be acquired using \textbf{t}:
\begin{equation}
\mathrm{t}=\operatorname{int}\left(\frac{\mathrm{abs}\left(\log _{2}(\mathrm{C})+\mathrm{b}\right)}{\gamma}\right) \, ,\
\end{equation}
where b and $\gamma$ are hyper-parameters set as 1 and 2 in our experiments, respectively.
If \textbf{t} is divisible by 2, \textbf{K} is equal to \textbf{t}, otherwise \textbf{K} is equal to \textbf{t} plus 1.
With the growing of channel depth, the EAC module can attain interaction among more channels. To limit the range of $\boldsymbol{C}$, sigmoid activation function $\sigma(\cdot)$ is applied to it. $\sigma(\cdot)$ can be expressed as:
\begin{equation}
\sigma(x)=\frac{1}{1+e^{-x}}.
\end{equation}
Then, we can get the final attention weights $D=\left[D_{1}, D_{2}, \ldots \ldots, D_{C}\right] \in \mathbb{R}^{1 \times C}$.
All the elements of $\boldsymbol{D}$ are in the range of 0 and 1. In other words, each element of $\boldsymbol{D}$ can be viewed as the key weight of the corresponding channel of the input feature map.
Finally, we perform an outer product of $ \boldsymbol{A}$ and $ \boldsymbol{D}$ to get the adjusted feature map $ \boldsymbol{E} \in \mathbb{R}^{H \times W \times C}$.
Thereby, RGB features and polarization features can be adjusted dynamically by the EAC module.
%Then, the adjusted features will be fused by the fusion module.

\textbf{Fusion module} is leveraged to fuse the adjusted feature maps from RGB branch and polarization branch following the EAC modules. 
   \begin{figure} [t]
   \begin{center}
   \begin{tabular}{c} %% tabular useful for creating an array of images 
   \includegraphics[height=2.5cm]{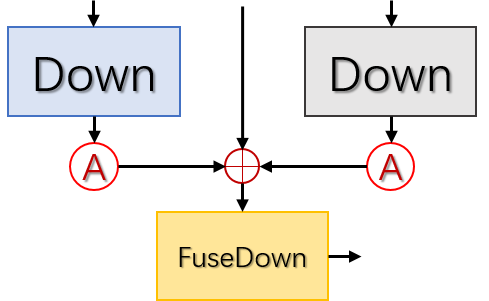}
   \end{tabular}
   \end{center}
   \vskip-4ex
   \caption[example] 
   { \label{fig:10}The structure of Fusion Module.}
   \vskip-2ex
   \end{figure}
As mentioned above, the fusion branch is the same as the RGB branch and the polarization branch.
The main difference lies in the inputting feature flow.
Assuming at the \textit{i}-th dowmsampling stage, the RGB branch's feature map is $y_{R G B}^{i} \in \mathbb{R}^{H_{i} \times W_{i} \times C_{i}}$,
and the polarization branch's feature map is $y_{P}^{i} \epsilon \mathbb{R}^{H_{i} \times W_{i} \times C_{i}}$.
Fig.~\ref{fig:10} illustrates one layer of the fusion branch for the fusion process.
The left part is the RGB feature, and the right part is the polarization feature, while the feature $m^{i}$ flowing through the center arrow is the fusion feature from the previous fusion stage. 
Then, the fused feature $m^{i+1}$ at the current stage can be expressed as:
\begin{equation}
m^{i+1}=y_{R G B}^{i} * E A C_{R G B}^{i}\left(y_{R G B}^{i}\right)+y_{P}^{i} * E A C_{P}^{i}\left(y_{P}^{i}\right)+m^{i}\, ,\
\end{equation}
where $m^{i+1}$ working as the fusion feature is passed into the fusion branch to extract higher-level features.
It should be noted that at the first fusion stage of our EAFNet architecture, it only has RGB feature and polarization feature as input information.

\section{Experiments and analysis}
In this section, the implementation details and a series of experiments with comprehensive analysis are presented. 
%The results showed a remarkable advancement with the aid of polarization fused SS, which conveys a brilliant future of multimodal SS.
\subsection{Implementation details}
The experiments concerning polarization fusion are performed on \textit{ZJU-RGB-P} dataset, while the preliminary experiment detailed in Section 1 and the supplement experiment detailed in Section 4.5 are performed on the \textit{Lost and Found} dataset~\cite{pinggera2016lost}.
The remaining implementation details are the same for all of the experiments.

For data augmentation, we first scale the images with random factors between 0.75 and 1.25, then we randomly crop the images with a crop size of 768$\times$768, followed with a random horizontal flipping.
It is worth noting that AoLP's random horizontal flipping has a critical difference from DoLP and RGB images.
According to Eq. (4), when the RGB images at four polarized directions are applied horizontal flipping, the AoLP will be:
\begin{equation}
A o L P^{'}=180^{\circ}-A o L P\, ,\
\end{equation}
where $A o L P^{'}$ is the ultimate horizontal flipped AoLP image, and $A o L P$ is merely the spatially horizontal flipped version of the initial AoLP image.
After all the data augmentation, all the processed images are normalized to the range between 0 and 1.

We use Tensorflow and an NVIDIA GeForce GTX 1080Ti GPU to implement EAFNet and perform training.
We use Adam optimizer~\cite{kingma2014adam} with an initial learning rate of 4$ \times $10$^{-4}$.
We decay the learning rate with cosine annealing to the minimum 2.5$ \times $10$^{-3}$ of the initial learning rate until the final epoch.
To combat over-fitting, we use the L2 weight regularization with a weight decay of 1$ \times $10$^{-4}$.
Unlike prior works~\cite{orvsic2019defense,sun2020real}, we have not adopted any pre-trained weights in order to investigate the effectiveness of multimodal SS fairly, with the aim to reach high performance even with limited pairs polarization images.
We utilize the cross entropy loss to train all the models with a batch size of 8.
We evaluate with the standard Intersection over Union (IoU) metric.

\subsection{Results and analysis} 
\begin{figure}[t] 
  \centering 
  \subfigure[DoLP distribution]{ 
    \includegraphics[height=4.4cm]{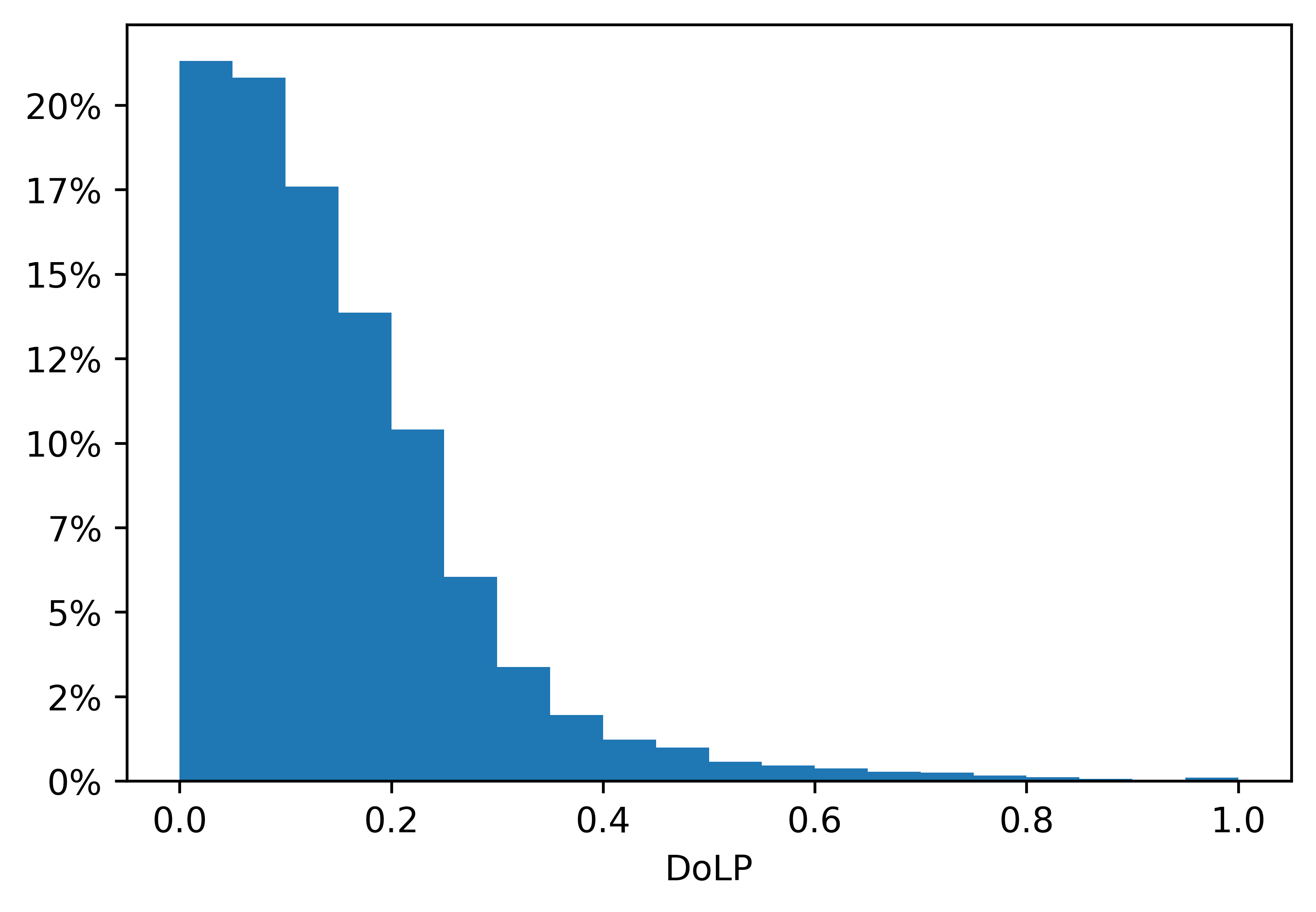}
  } 
  \subfigure[AoLP distribution]{ 
    \includegraphics[height=4.4cm]{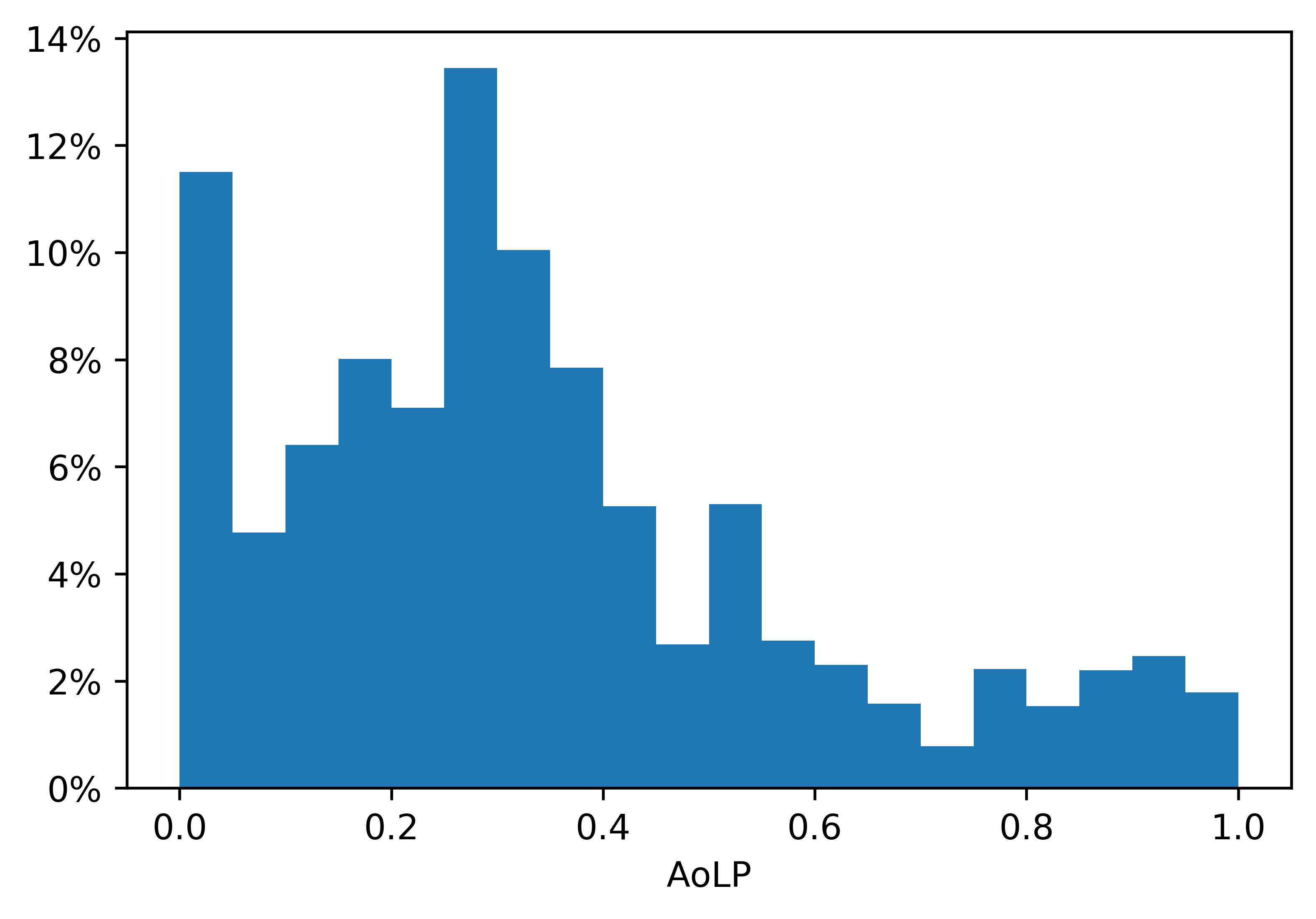} 
  } 
  \vskip-2ex
  \caption{\label{fig:11}The value distribution of \textit{ZJU-RGB-P} training set's DoLP and AoLP: (a) The DoLP distribution; (b) The AoLP distribution. All the values are normalized to the range between 0 and 1.} 
  \vskip-2ex
\end{figure} 
Both of AoLP and DoLP can represent polarization information of scenes, but which is the better to be fused into polarization-driven SS remains an open question.

We have made a brief analysis of the superiority of AoLP over DoLP for polarization-driven SS in Section 3.1 intuitively. 
As preliminary investigation, according to Fig.~\ref{fig:4}, we find that AoLP's distribution has a remarkable difference to that of DoLP.
Further, we present the statistics of the value distributions of DoLP and AoLP on the \textit{ZJU-RGB-P} training set, as shown in Fig.~\ref{fig:11}. 
The majority of all pixels of the training set are with a small DoLP ranging from 0 to 0.4, while the portion of pixels whose DoLP values are larger than 0.4 is rather low, indicating that DoLP offers limited information, merely on categories with highly polarized characteristics.
Different from DoLP, AoLP offers a uniform distribution.
%other than the value near 0.
In other words, nearly all pixels of AoLP images possess meaningful features that are useful for SS.
With the different distributions of DoLP and AoLP, it can be expected that their information are emphasized on different polarization characteristics.
Inspired by this observation, we perform a series of experiments to investigate the effectiveness of EAFNet to fuse polarization and RGB information, and whether AoLP is superior to DoLP in offering complementary information for RGB-P image segmentation.

For the basic control experiment, we first train the RGB-only SwiftNet on \textit{ZJU-RGB-P} dataset as our \textbf{Baseline}.
Then, four sets of training are performed for comparison.
As shown in Fig.~\ref{fig:8}, EAFNet is a two-path network, where RGB images and polarization information are input into different paths.
To explore the better polarization feature, we select AoLP (marked as \textbf{EAF-A}) and DoLP (marked as \textbf{EAF-D}) as the input polarization information, respectively.
Considering the fact that both AoLP and DoLP can offer polarization information, we also concatenate AoLP and DoLP images along channel to build a polarization representation for training a variant model of EAFNet (marked as \textbf{EAF-A/D}). 
\begin{table}[t]
\caption{\textbf{Accuracy analysis on \textit{ZJU-RGB-P}} including per-class accuracy in IoU (\%).}
\vskip-2ex
\label{tab:1}
\setlength{\tabcolsep}{1.3mm}{
\begin{center}
\begin{tabular}{|c|c|c|c|c|c|c|c|c|c|}
\hline
\multicolumn{1}{|c|}{\textbf{Model}}  & Building      & Glass         & Car           & Road          & Vegetation    & Sky           & Pedestrian    & Bicycle       & mIoU          \\
\hline
\rule[-1ex]{0pt}{3.5ex} \textbf{Baseline}  & 83.0          & 73.4          & 91.6          & 96.7          & 94.5          & 84.7          & 36.1          & 82.5          & 80.3          \\
\hline
\rule[-1ex]{0pt}{3.5ex} \textbf{EAF-A} & \textbf{87.0} & \textbf{79.3} & \textbf{93.6} & \textbf{97.4} & \textbf{95.3} & \textbf{87.1} & 60.4 & 85.6 & \textbf{85.7} \\
\hline
\rule[-1ex]{0pt}{3.5ex} \textbf{EAF-D} & 86.4 & 76.9 & 93.0 & 97.1 & \textbf{95.5} & 86.1 & 62.1 & \textbf{86.0} & 85.4 \\
\hline
\rule[-1ex]{0pt}{3.5ex} \textbf{EAF-A/D} & \textbf{87.1} & 77.7 & \textbf{93.7} & \textbf{97.4} & 95.2 & 84.9 & \textbf{63.8} & 83.9 & 85.4 \\
\hline
\rule[-1ex]{0pt}{3.5ex} \textbf{EAF-3Path} & 83.8 & 74.4 & 92.0 & 95.9          & 95.1 & 86.5 & 59.2 & 80.2          & 83.4 \\ 
\hline
\end{tabular}
\end{center}}
\vskip-2ex
\end{table}
Finally, we build a three-path version of EAFNet, where we select one path as the RGB path and the other two paths as polarization paths. AoLP and DoLP are passed into the two polarization paths (marked as \textbf{ EAF-3Path}).

All quantitative results of the experiments are shown in Table~\ref{tab:1}. 
It can be seen that models combined with polarization information can advance the segmentation of objects with polarization characteristics like glass (73.4\% to 79.3\%), car (91.6\% to 93.7\%) and bicycle (82.5\% to 86.0\%). 
In addition, we observe that not only the IoU of classes with polarization characteristics are advanced, but other classes' IoU have been improved by a great extent when combined with polarization information, especially pedestrian (36.1\% to 63.8\%).
Meanwhile, the mIoU is lifted to 85.7\% from 80.3\%.

Further, we compare and discuss among the groups that fuse polarization-based features.
As shown in Table~\ref{tab:1}, \textbf{EAF-A} is the optimal setting, while \textbf{EAF-3Path} is the worst group.
Here, we analyze from the view of data distribution and model complexity.
Our main focus is on the classes like glass and car, as the initial motivation of this study is to lift the segmentation performance of objects with polarization characteristics.
Making a comparison between \textbf{EAF-A} and \textbf{EAF-D}, we find that the former can better advance the IoU of glass and car than the latter. 
It is the data distribution that counts.
The analysis of the different distributions in previous sections shows that AoLP offers a better spatial representation like contour information than DoLP, while DoLP only offers meaningful information on areas with high polarization information.
In this sense, AoLP provides richer priors and complementary information for RGB-P segmentation.
The reason why \textbf{EAF-A/D} can attain higher IoU values on glass and car than \textbf{EAF-D} is that AoLP complements the spatial features of DoLP.
However, it reaches a lower IoU on glass than \textbf{EAF-A}, because the interference between DoLP and AoLP features occurs, bringing some side effects and losing some serviceable information.
\textbf{EAF-3Path} is the worst group as the complex architecture of this model prevents the model from exploiting the most informative features.
Besides, the 3-path structure impairs the capacity of RGB features, which is critical for outdoor scene perception.
Eventually, we conclude from the quantitative analysis that utilizing AoLP images to feed in the polarization path can greatly advance polarization-driven segmentation performance.

\begin{figure}[t] 
  \centering 
  \subfigure[RGB Image]{ 
    \includegraphics[height=6cm]{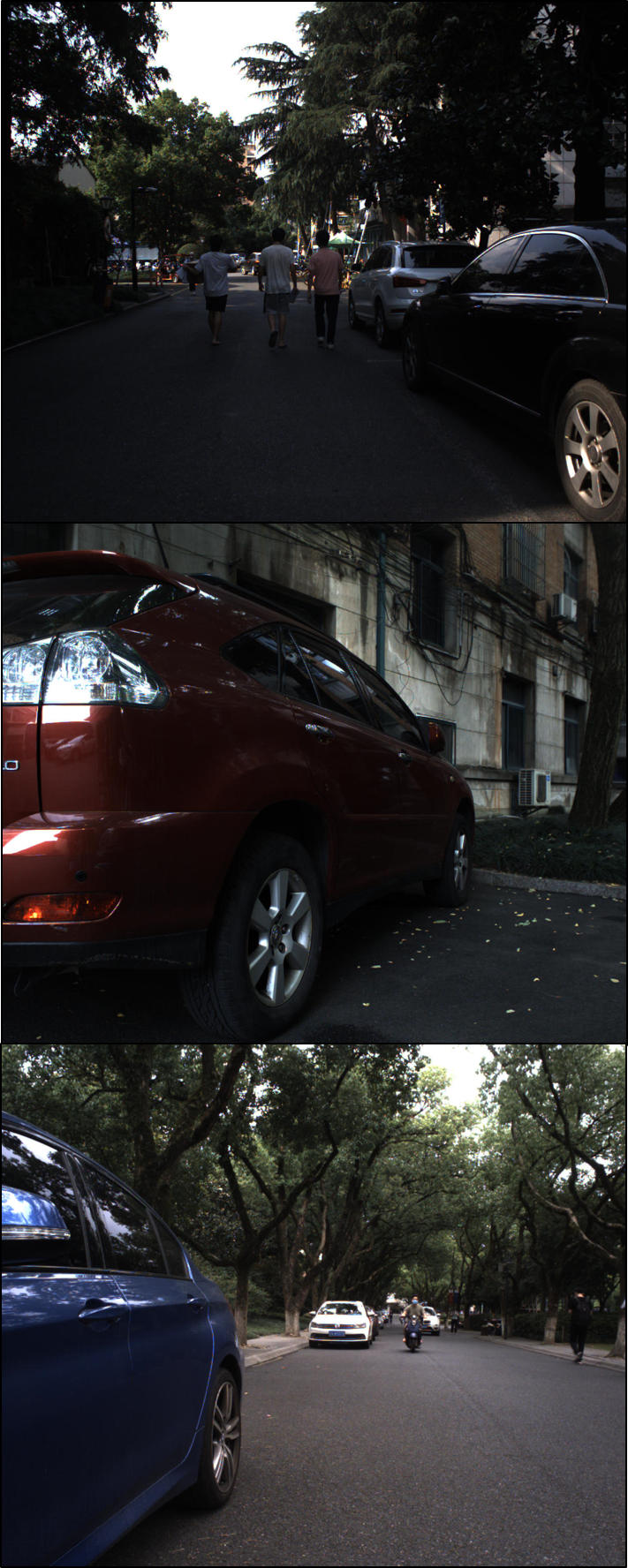}
  } 
    \subfigure[AoLP]{ 
    \includegraphics[height=6cm]{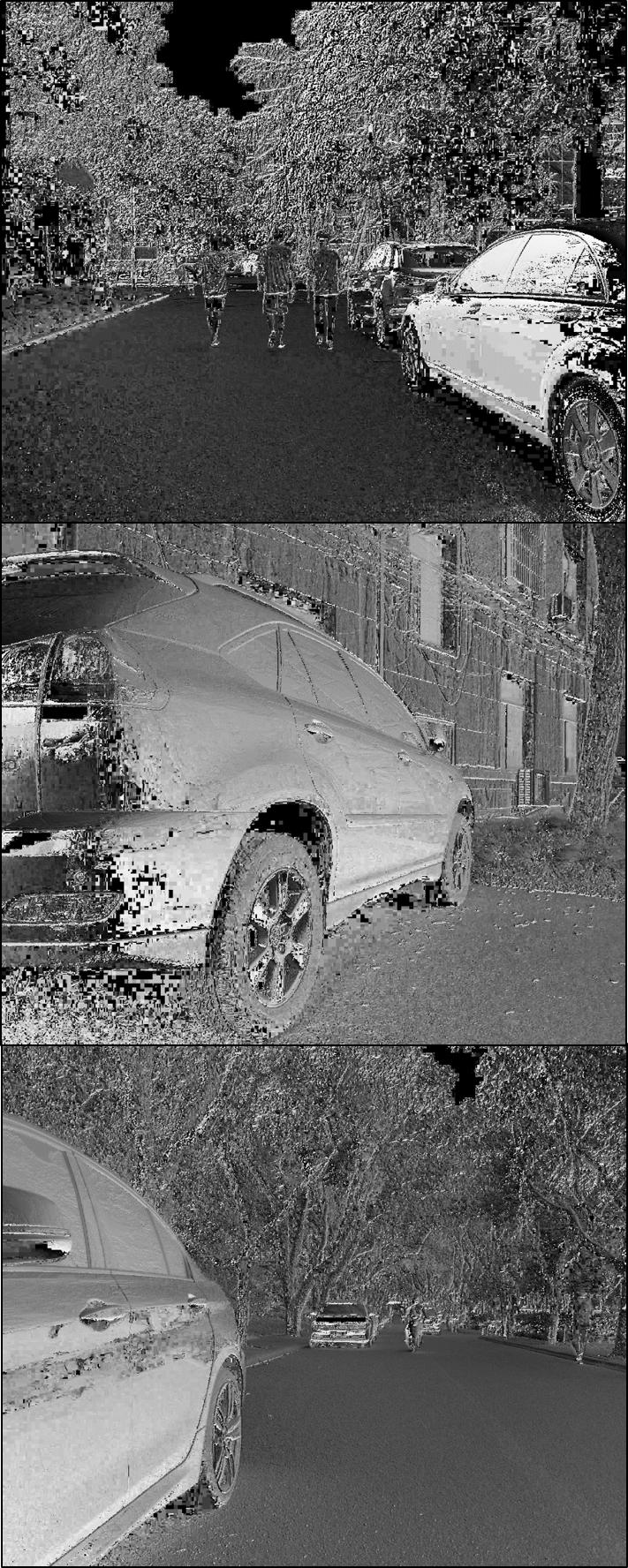} 
  } 
  \subfigure[Label]{ 
    \includegraphics[height=6cm]{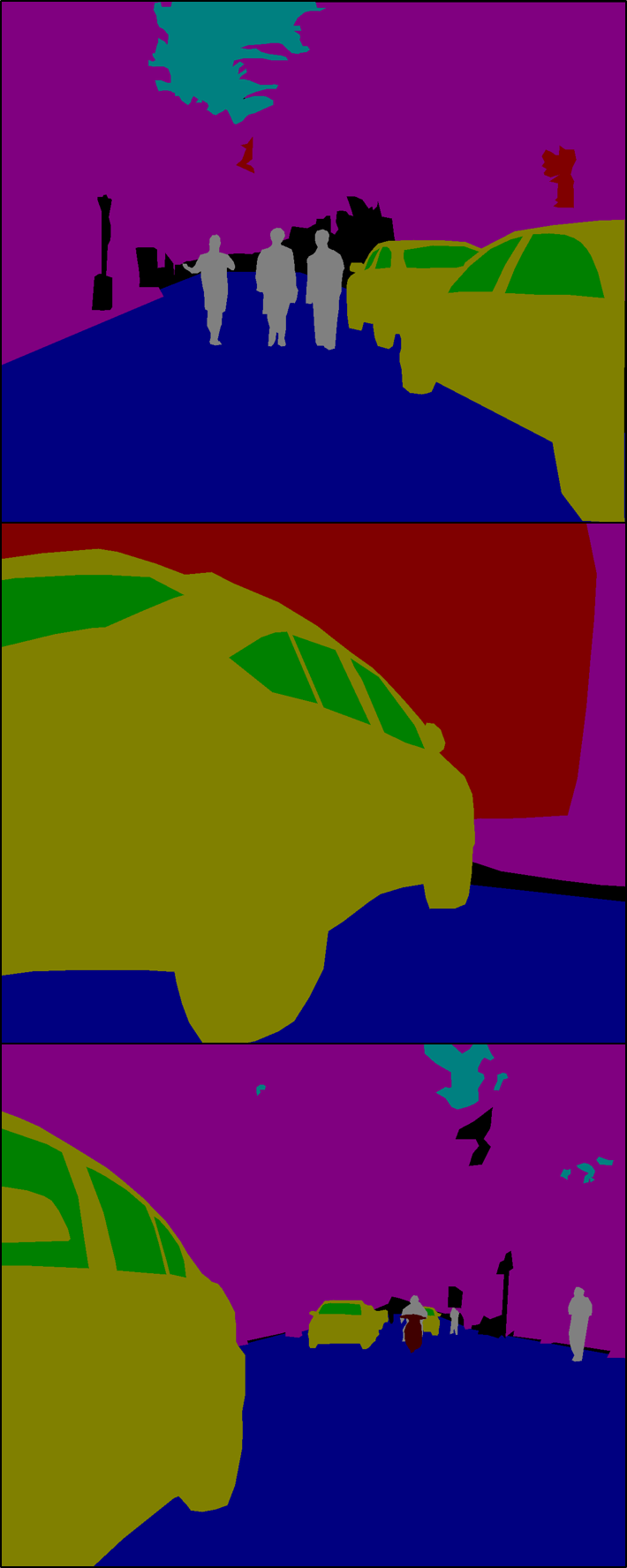} 
  } 
  \subfigure[Baseline]{ 
    \includegraphics[height=6cm]{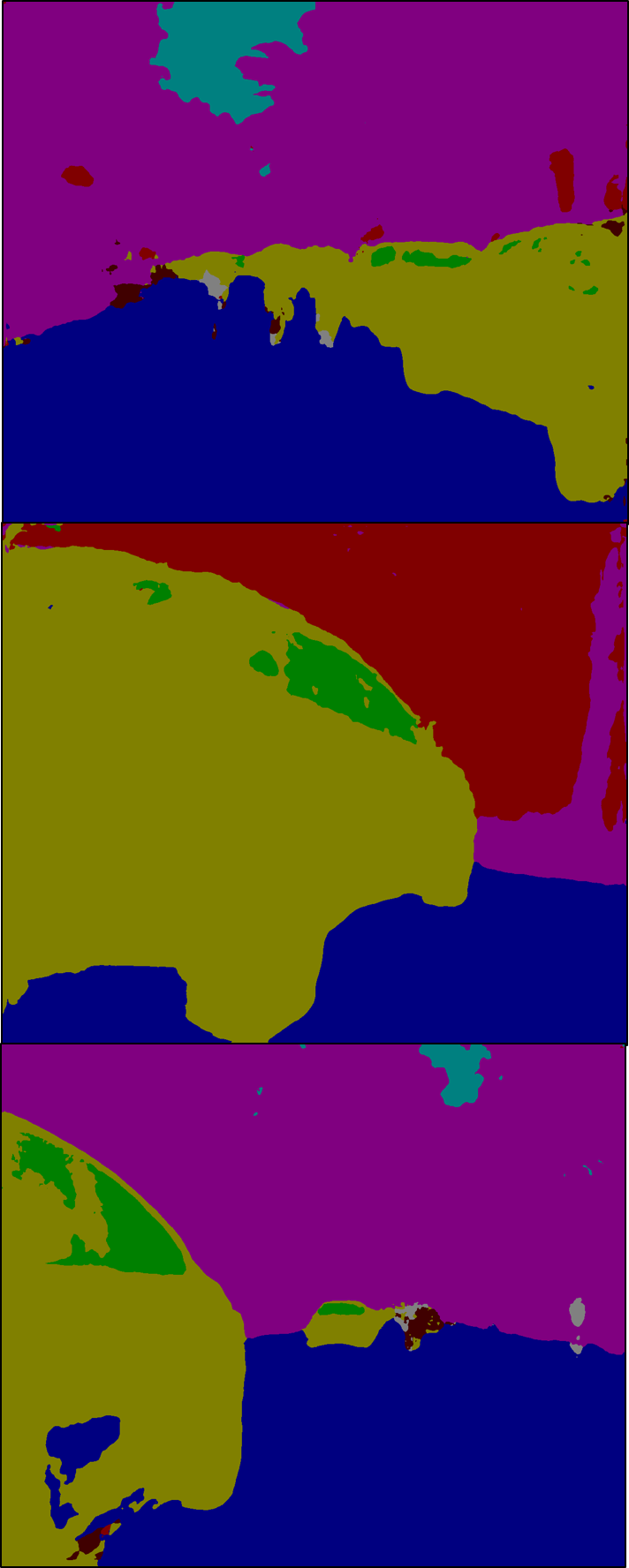} 
  } 
  \subfigure[EAFNet]{ 
    \includegraphics[height=6cm]{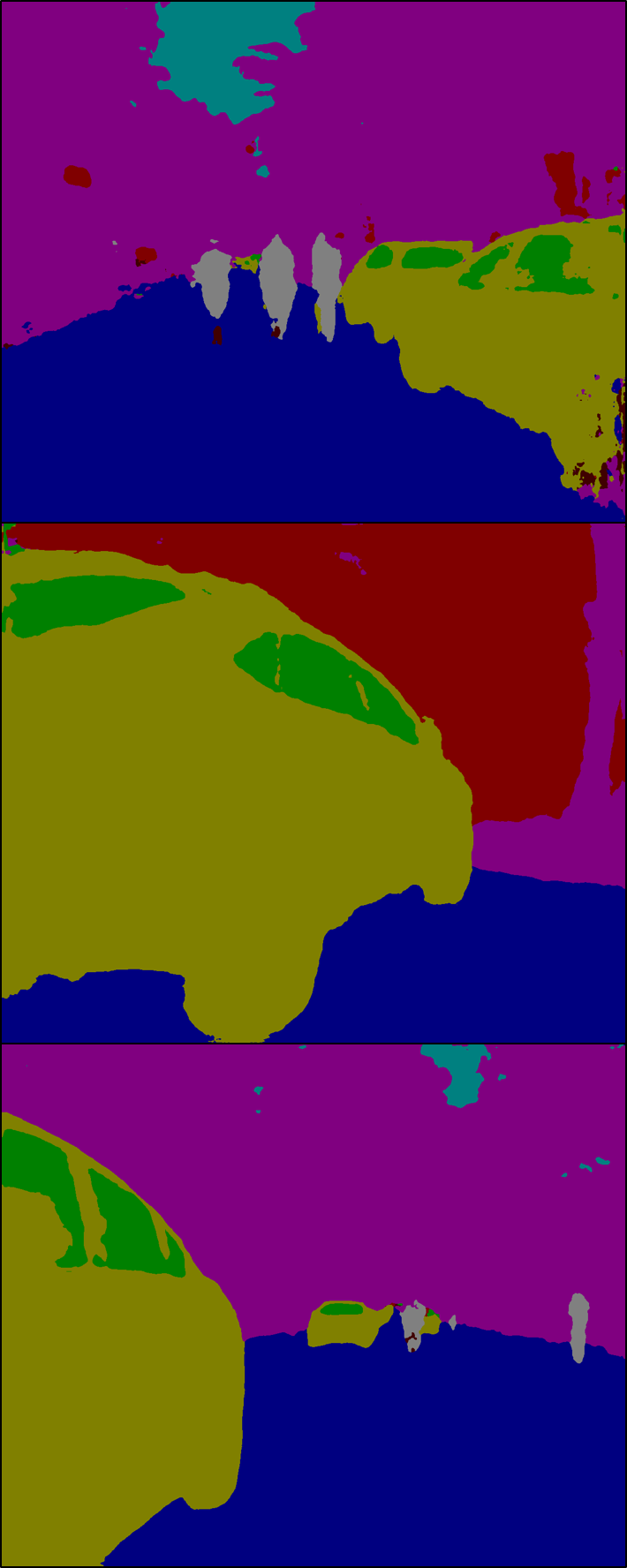} 
  } 
  \vskip-2ex
  \caption{\label{fig:12}Qualitative result comparison between the RGB-only baseline and our EAFNet.} 
  \vskip-2ex
\end{figure} 
For qualitative results, we use the \textbf{Baseline} RGB-only model and the \textbf{EAF-A} polarization-driven model, \textit{i.e.}, SwiftNet and our EAFNet fed with AoLP on the \textit{ZJU-RGB-P} validation set to produce a series of visualization examples, as illustrated in Fig.~\ref{fig:12}. 
We find that SwiftNet wrongly segments the pedestrians into cars in the first row of Fig.~\ref{fig:12} where EAFNet detects them correctly.
In the second row, EAFNet successfully distinguishes glass from the car, while SwiftNet can not segment the full glass area.
Moreover, SwiftNet even segments part of the car into road and part of the pedestrian into vegetation according to the last row of results in Fig.~\ref{fig:12}.
The wrong segmentation results in outdoor traffic scenes can lead to terrible situations and even accidents once the model is selected to guide autonomous vehicles or assisted navigation~\cite{zhang2020issafe,yang2019robustifying}.
\CHANGE{In addition, we can learn from the first row that EAFNet can keep a good performance in low luminance. The area to the fore are in the shade with low luminance, where limited RGB information makes it hard for our eyes to perceive the scene, while AoLP offers sufficient spatial information to complement it, thereby advancing the performance of EAFNet. On the contrary, the area in the distance is illuminated by sun with high intensity annotated as background, where the complementary AoLP aids the segmentation of EAFNet, leading to accurate perception in such challenging scenarios.}
It is obvious that polarization-driven SS can complement the missing information merely based on RGB images. Therefore, multimodal SS is beneficial for semantic understanding in pursuit of robust outdoor scene perception.

\subsection{Analysis of EAC module}
   \begin{figure} [t]
   \begin{center}
   \begin{tabular}{c} %% tabular useful for creating an array of images 
   \includegraphics[height=6.7cm]{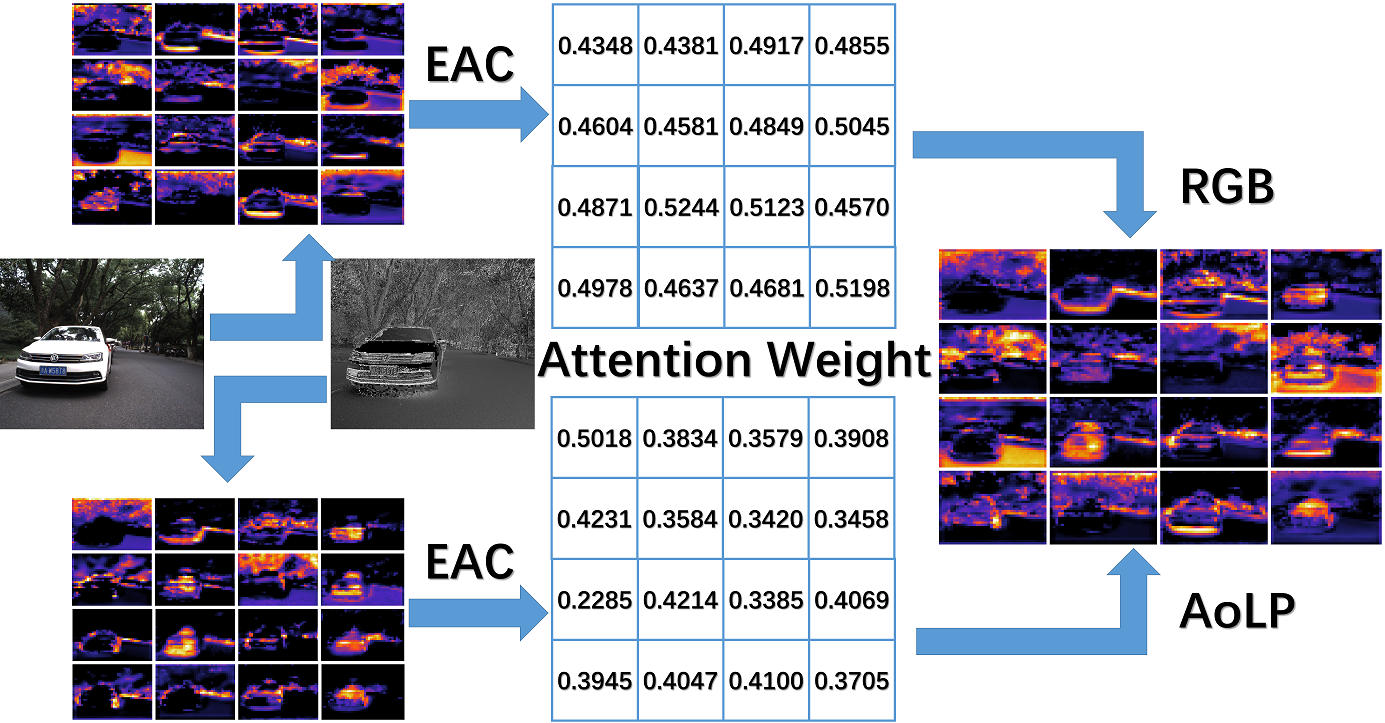}
   \end{tabular}
   \end{center}
    \vskip-4ex
   \caption[example] 
   { \label{fig:13}The EAC module's role in fusing features: An RGB image and an AoLP image are fed into EAFNet, then we visualize the fourth downsampling block's feature maps. Following that, EAC module extracts RGB and polarization branches' attention weights. Finally, the feature maps are adjusted by the attention weights and fused.}
    %\vskip-2ex
   \end{figure}
   
   \begin{figure} [t]
   \begin{center}
   \begin{tabular}{c} %% tabular useful for creating an array of images 
   \includegraphics[height=3cm]{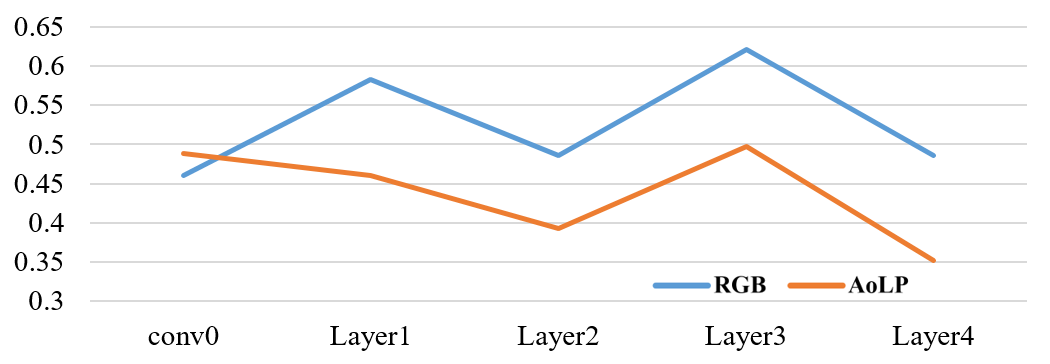}
   \end{tabular}
   \end{center}
   \vskip-4ex
   \caption[example] 
   { \label{fig:14}The average attention weights of EAFNet at all levels.}
   \vskip-2ex
   \end{figure}
EAC module is the key module of EAFNet, which can extract attention weights of RGB branch and AoLP branch.
To better demonstrate the effect of EAC module, we visualize the fourth downsampling block's feature maps of RGB and AoLP branches, and their attention weights of EAC module as shown in Fig.~\ref{fig:13}.
We only visualize feature maps of the former 16 channels.
Here, \textbf{(i, j)} denotes the position at the i-th row and the j-th col of the feature map, which corresponds to the attention weight one by one, where some insightful results can be found.
In the RGB branch, we find that the car and glass area 
have low responses in the feature maps.
On the contrary, the corresponding area of the AoLP branch have high responses, especially at \textbf{(2, 4)}, \textbf{(3, 2)} and \textbf{(4, 4)}.
Then, their EAC module extracts their attention weights, respectively. Taking \textbf{(3, 2)} as an example to illustrate the complement process, this channel's attention weights are 0.5244 and 0.4214 for RGB and AoLP branches, respectively.
Then, the corresponding feature map will be multiplied by the attention weight.
Finally, the adjusted feature map are added up to build the ultimate feature map, and it can be clearly seen that the feature maps spotlight the area of car perfectly.

As it can be seen in Fig.~\ref{fig:13}, the attention weights of RGB are higher than those of AoLP in most cases.
Here, we evaluate the weights generated by EAC Module at all levels and illustrate the average of them as shown in Fig.~\ref{fig:14}. 
According to the curve, it can be easily observed that the RGB branch possesses higher weights than the AoLP branch at Layer1, Layer2, Layer3 and Layer4. 
On the contrary, the AoLP branch has a higher weight than the RGB branch at the first downsampling block, \textit{i.e.}, conv0.
As mentioned before, AoLP offers a representation of spatial information and rich priors. 
Therefore, at the beginning of EAFNet, AoLP offers more distinguished features than RGB. 
With the features flowing into deeper layers, RGB branch becomes overwhelming.
In addition, both of the weight curves keep a similar variation trend and reach the highest at Layer3.

\subsection{Ablation study}
\begin{table}[t]
\caption{\textbf{Accuracy analysis of the ablation study on \textit{ZJU-RGB-P}} (\%).}
\vskip-2ex
\label{tab:2}
\setlength{\tabcolsep}{1.3mm}{
\begin{center}
\begin{tabular}{|c|c|c|c|c|c|c|c|c|c|}
\hline
\multicolumn{1}{|c|}{\textbf{Model}}  & Building      & Glass         & Car           & Road          & Vegetation    & Sky           & Pedestrian    & Bicycle       & mIoU          \\
\hline
\rule[-1ex]{0pt}{3.5ex} \textbf{Baseline}  & 83.0          & 73.4          & 91.6          & 96.7          & 94.5          & 84.7          & 36.1          & 82.5          & 80.3          \\
\hline
\rule[-1ex]{0pt}{3.5ex} \textbf{SN-A} & 74.0 & 66.6 & 87.1 & 94.7 & 91.1 & 76.1 & 32.9 & 65.5 & 73.5 \\
\hline
\rule[-1ex]{0pt}{3.5ex} \textbf{SN-RGB/A} & 83.9 & 75.6 & 91.6 & 96.9 & 94.4 & 78.2 & 41.0 & 79.9 & 80.2 \\
\hline
\rule[-1ex]{0pt}{3.5ex} \textbf{EAF-wo-A} & 85.2 & 75.4 & 92.2 & 96.8 & 94.8 & 84.4 & 40.6 & 82.9 & 81.6 \\
\hline
\rule[-1ex]{0pt}{3.5ex} \textbf{EAF-A} & \textbf{87.0} & \textbf{79.3} & \textbf{93.6} & \textbf{97.4} & \textbf{95.3} & \textbf{87.1} & \textbf{60.4} & \textbf{85.6} & \textbf{85.7} \\
\hline
\end{tabular}
\end{center}}
%\vskip-2ex
\end{table}

To better illustrate that fusing polarization information is beneficial and to verify \YKL{EAFNet's powerful fusion capacity,}
% that EAFNet has the better capacity to fuse features than a simple concatenation of RGB and polarization images, 
we have performed \YKL{three} extra training.
First, we directly utilize AoLP images to train SwiftNet, denoted as \textbf{SN-AoLP}.
Second, we utilize the concatenation of RGB and AoLP images to train SwiftNet, denoted as \textbf{SN-RGB/A}.
\YKL{Finally, we get rid of the EAC Moudle of EAFNet, and element-wise addition is utilized to fuse RGB features and AoLP features instead, denoted as \textbf{EAF-wo-A}.}
We also select \textbf{Baseline} and \textbf{EAF-A} of the previous experiment for the ablation study, as shown in Table~\ref{tab:2}.

We find that \textbf{SN-AoLP} can reach a decent performance, which is benefited from the spatial priors of AoLP.
It can be learned from the comparison between \textbf{Basline} and \textbf{SN-AoLP} that RGB possesses more distinguished features than AoLP, while AoLP offers meaningful information as well.
Besides, \textbf{SN-RGB/A} can indeed advance the segmentation of classes with polarization characteristics like glass (73.4\% to 75.6\%), but it does not yield remarkable benefits for all classes in contrast \YKL{to} our \textbf{EAF-A}.
In addition, \textbf{SN-RGB/A} even causes a slight mIoU degradation to 80.2\% from 80.3\%.
It is the difference between RGB and AoLP distributions that accounts for this degradation.
The interference between RGB and AoLP has an adverse impact on the extraction of distinguishable features.
\YKL{We can learn from the comparison between \textbf{EAF-wo-A} and \textbf{EAF-A} that attention mechanism of EAFNet is highly effective. Combining with EAC Module, all classes have a remarkable elevation like glass (75.4\% to 79.3\%), pedestrian (40.6\% to 60.4\%) and so on.}
Making a comparison among all the groups, we draw a conclusion that \textbf{EAF-A} reaches the highest accuracy on all classes, indicating the effectiveness of our EAFNet and the designed polarization fusion strategy. 

\subsection{Generalization to other sensor fusion}
To prove the flexibility of EAFNet to be adapted to other sensor combination scenarios besides polarization information, we utilize disparity images with EAFNet to investigate in the unexpected obstacle detection scenario, \textit{i.e.}, the preliminary investigation mentioned in Section 1.

According to Fig.~\ref{fig:1}, we can find that disparity images reflect the contours of the tiny obstacles.
Thereby, combing RGB and disparity images with EAFNet is hopeful to address the devastating results of using merely RGB data.
Considering the similar distribution between disparity and AoLP images, we train the EAFNet fed with pixel-aligned disparity and RGB images (marked as \textbf{EAFNet-RGBD}). 
\begin{table}[t]
\caption{\textbf{Accuracy analysis of the supplement experiment on \textit{Lost and Found}} (\%).}
\vskip-2ex
\label{tab:3}
\begin{center}
\begin{tabular}{|c|c|c|c|c|}
\hline 
\multicolumn{1}{|c|}{\textbf{Group} } & \multicolumn{2}{c|}{\textbf{SwiftNet-RGB}} &\multicolumn{2}{c|}{\textbf{EAFNet-RGBD}}\\
\hline
\multicolumn{1}{|c|}{\textbf{Class}} & Road & Obstacle & Road & Obstacle\\
\hline
\rule[-1ex]{0pt}{3.5ex} \textbf{Precision} & 85.4& 26.5&\textbf{88.2}& \textbf{76.2}\\
\hline
\rule[-1ex]{0pt}{3.5ex} \textbf{Recall} & 63.9&49.8& \textbf{75.9}& \textbf{63.0}\\
\hline
\rule[-1ex]{0pt}{3.5ex} \textbf{IoU} & 56.2& 20.9& \textbf{68.9}& \textbf{52.7}\\
\hline
\end{tabular}
\end{center}
\vskip-2ex
\end{table}
\begin{figure}[t] 
  \centering 
  \subfigure[RGB Image]{ 
    \includegraphics[height=1.5cm]{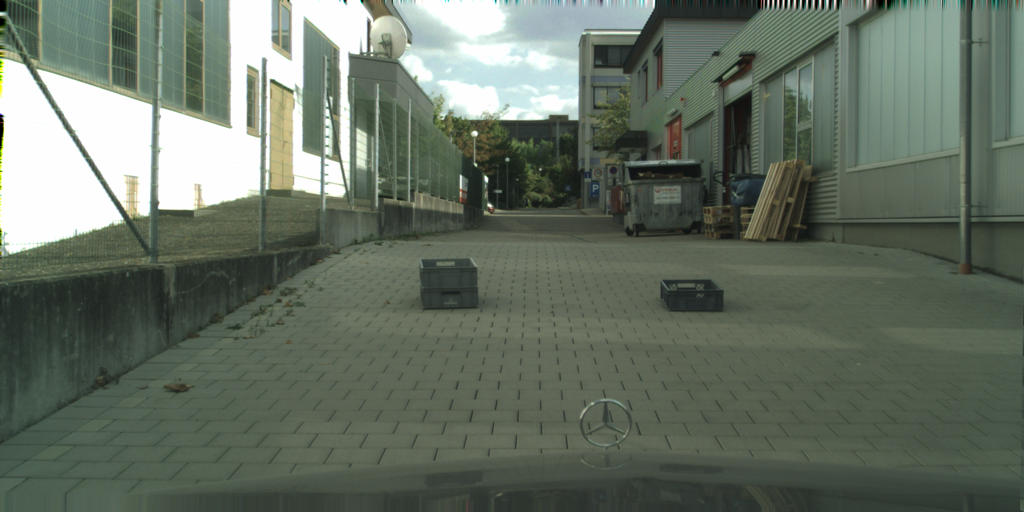}
  } 
  \subfigure[Label]{ 
    \includegraphics[height=1.5cm]{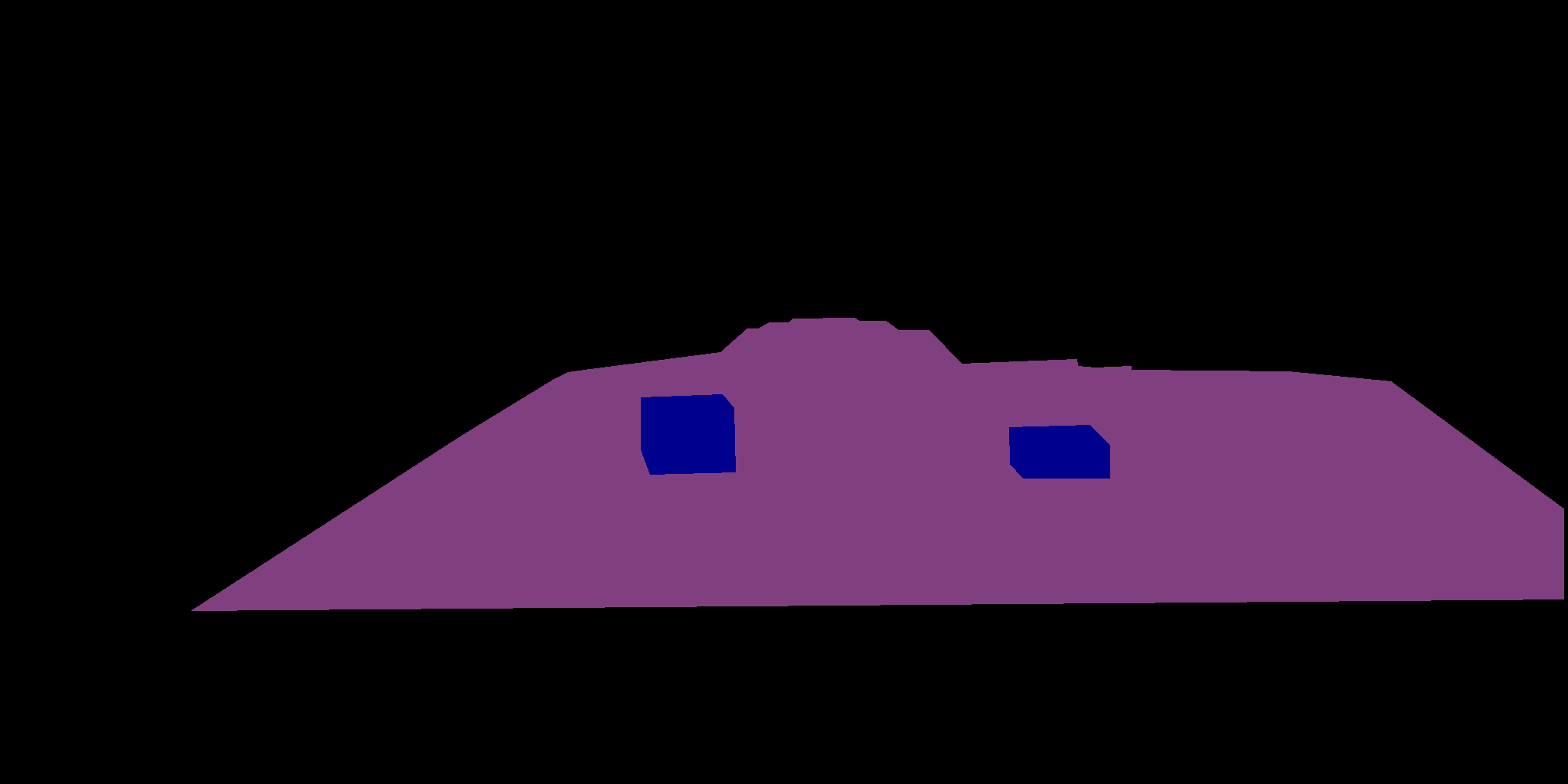} 
  } 
    \subfigure[SwiftNet-RGB]{ 
    \includegraphics[height=1.5cm]{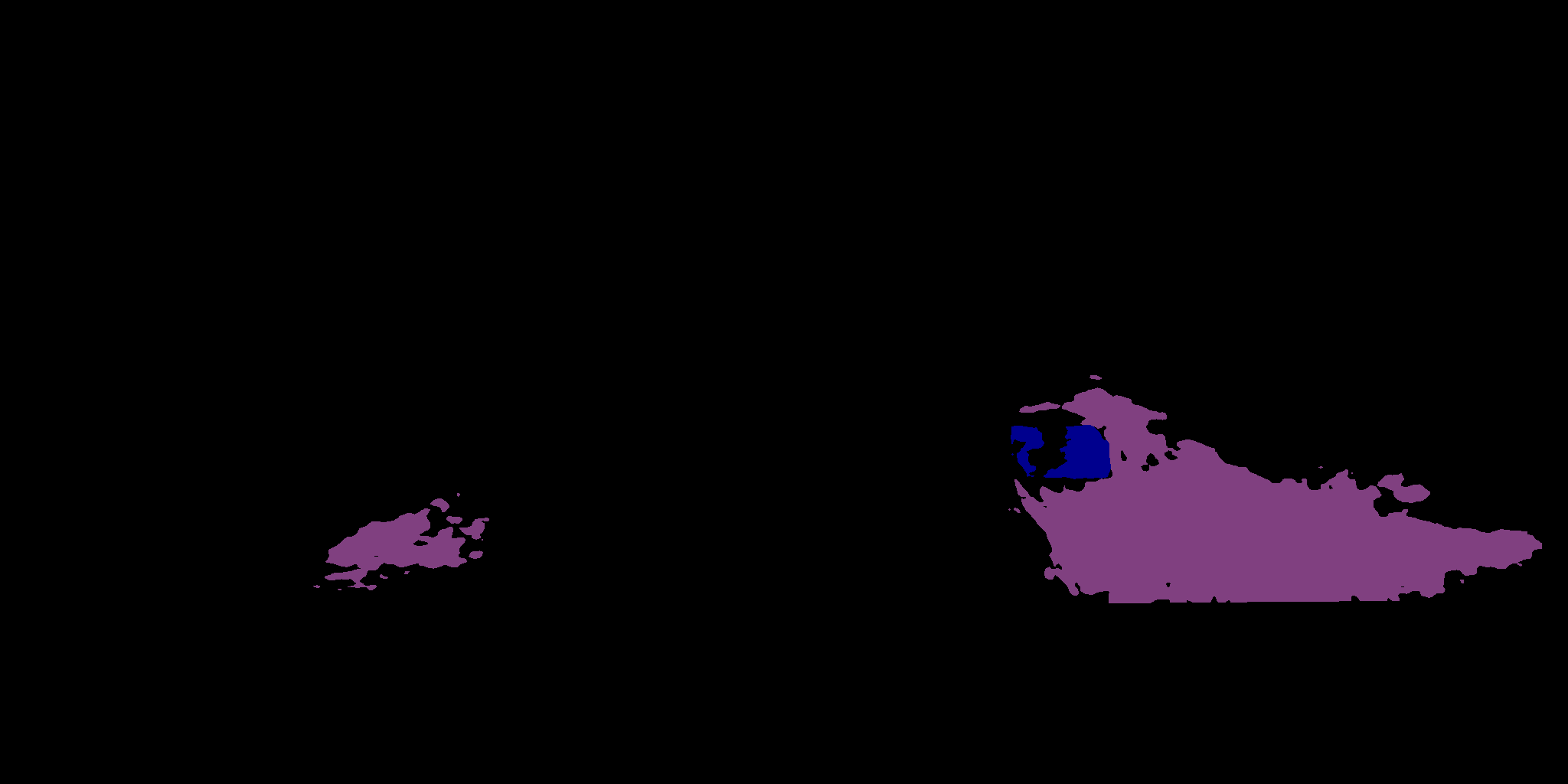} 
  } 
    \subfigure[EAFNet-RGBD]{ 
    \includegraphics[height=1.5cm]{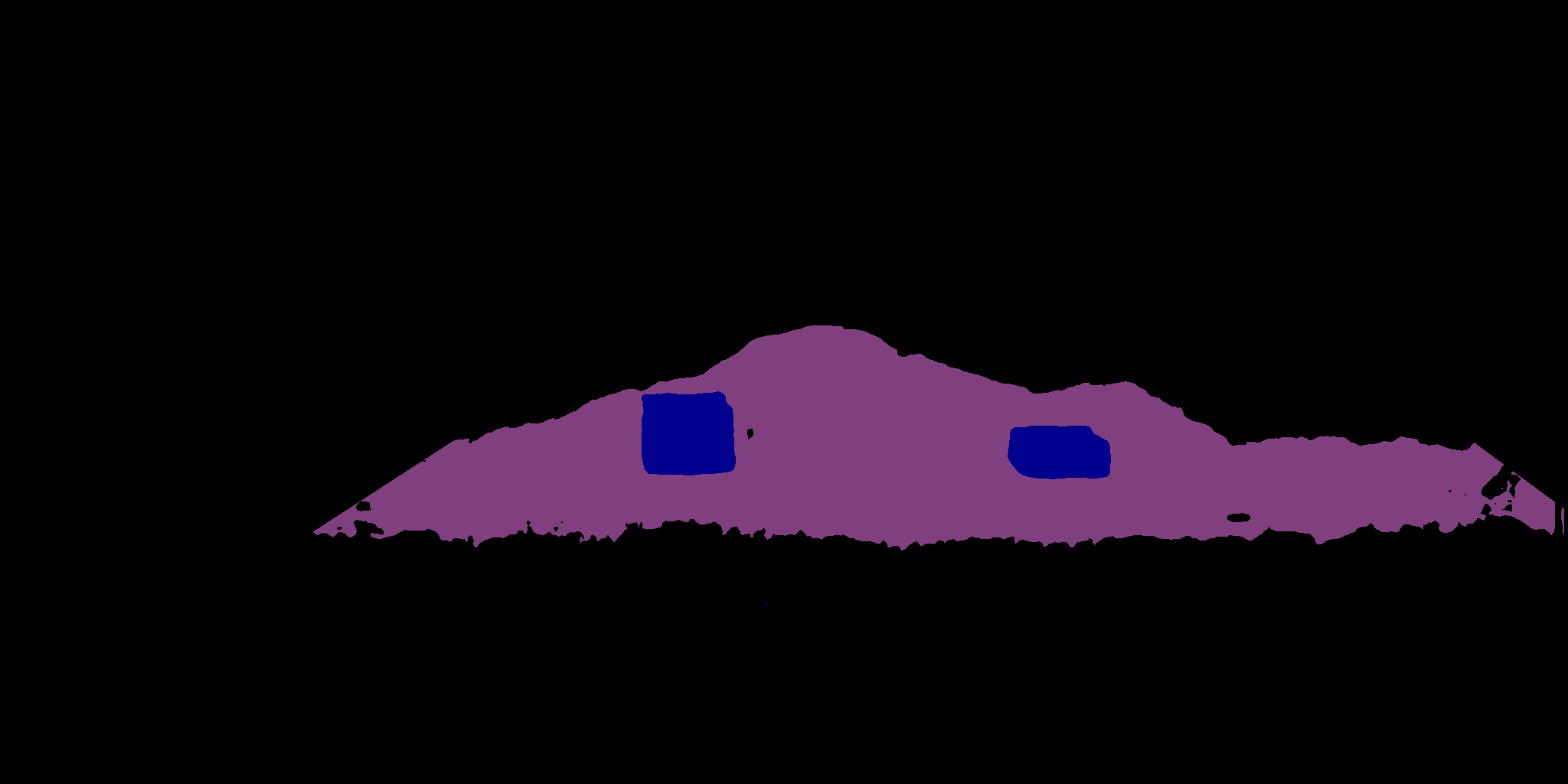} 
  } 
  \vskip-2ex
  \caption{\label{fig:15}Qualitative comparison results of SwiftNet-RGB and EAFNet-RGBD.}
  \vskip-2ex
\end{figure} 
We mark the baseline group as \textbf{SwiftNet-RGB}.
All the training strategies are set according to Section 4.1.
From the results in Table~\ref{tab:3}, we observe a remarkable elevation on the performance with the aid of EAFNet and complementary disparity images,
where the precision and IoU of obstacle segmentation are lift to 76.2\% from 26.5\% and 52.7\% from 20.9\%, respectively.
It indicates that combing disparity images with EAFNet bears fruit.
To have a better realization of \textbf{EAFNet-RGBD}'s effects intuitively, we visualize an example as shown in Fig.~\ref{fig:15}, where \textbf{EAFNet-RGBD} segments obstacles and most of the road areas successfully, but \textbf{SwiftNet-RGB} ignores most of the road and obstacles.
Therefore, it is essential to perform multimodal semantic segmentation with complementary sensing information like polarization-driven and depth-aware features to have a reliable and holistic understanding of outdoor traffic scenes.

\section{Conclusion and future work}
In this paper, we propose EAFNet by fusing features of RGB and polarization images. We build \textit{ZJU-RGB-P} with our integrated multimodal vision sensor, which is the first RGB-polarization semantic segmentation dateset to the best of our knowledge.
EAFNet dynamically extracts attention weights of RGB and polarization branches, adjusts and fuses multimodal features, significantly advancing the segmentation performance, especially on classes with highly polarized characteristics like glass and car.
Extensive experiments are conducted to prove the effectiveness of EAFNet for incorporating features from different sensing modalities and the flexibility to be adapted to other sensor combination scenarios like RGB-D perception.
Therefore, EAFNet is a multimodal SS model that can be utilized in diverse real-world applications.

In the future, there are two research paths that can be explored.
One is to build more kinds of multimodal dataset based on the integrated multimodal vision sensor like RGB-Infrared dataset to address nighttime scene understanding.
Another line is to enlarge the categories of \textit{ZJU-RGB-P} to cope with the detection of transparent objects, ice and water hazards.

\section*{Funding}
This research was granted from ZJU-Sunny Photonics Innovation Center (No. 2020-03). This research was also funded in part through the AccessibleMaps project by the Federal Ministry of Labor and Social Affairs (BMAS) under the Grant No. 01KM151112.

\section*{Acknowledgments}
This research was supported in part by Hangzhou SurImage Technology Company Ltd.

\section*{Disclosures}
The authors declare that there are no conflicts of interest related to this article.
%%The authors declare no conflicts of interest.

\bibliography{sample}

\end{document}